\documentclass[10pt,twoside]{article}
\usepackage{amsmath}
\usepackage[utf8]{inputenc}
\usepackage{url}
\usepackage{dsfont}
\usepackage{graphicx}
\usepackage{yhmath}
\usepackage[table]{xcolor}
\usepackage{mathrsfs} % permet d'avoir le D cursive
\usepackage{amssymb}
\usepackage{url}
\usepackage{makecell}
\usepackage{arydshln}
\usepackage{multirow}
\usepackage{arydshln}
\usepackage{mathtools}
\usepackage{stmaryrd}  % pour intervalle entiers
 \usepackage{amsthm,amssymb}
\usepackage{hyperref}

\input epsf
\def\limiten{\renewcommand{\arraystretch}{0.5}
\begin{array}[t]{c}\stackrel{}{\longrightarrow} \\
{\scriptstyle n\rightarrow
\infty}\end{array}\renewcommand{\arraystretch}{1}}

\numberwithin{equation}{section}

\newtheorem{thm}{Theorem}[section]

\newtheorem{Corol}[thm]{Corollary}
\newtheorem{Def}[thm]{Definition\rm}
\newtheorem{example}[thm]{Exemple}

\newtheorem{lem}[thm]{Lemma}

\newtheorem{rmrk}[thm]{Remark}

\newcommand{\E}{\ensuremath{\mathbb{E}}}
\newcommand{\R}{\ensuremath{\mathbb{R}}}
\newcommand{\Z}{\ensuremath{\mathbb{Z}}}

\newcommand{\N}{\ensuremath{\mathbb{N}}}

\newcommand{\var}{\ensuremath{\mathrm{Var}}}

\newcommand{\lip}{\ensuremath{\mathrm{Lip}}}
\definecolor{grisclair}{gray}{0.9}
\font\dsrom=dsrom10 scaled 1200
\def \ind{\textrm{\dsrom{1}}}
\DeclareMathOperator*{\argmin}{argmin}

 \newcommand{\mk}{ { \mathcal{K}} }
\newcommand{\mx}{ { \mathcal{X}} }

\renewcommand{\arraystretch}{.8}

\begin{document}
%\date{ }

\title{\bf Deep learning from strongly mixing observations: Sparse-penalized regularization and minimax optimality}
 \maketitle \vspace{-1.0cm}

\begin{center}
      William Kengne
     % \footnote{Developed within the ANR BREAKRISK: ANR-17-CE26-0001-01 and the  CY Initiative of Excellence (grant "Investissements d'Avenir" ANR-16-IDEX-0008), Project "EcoDep" PSI-AAP2020-0000000013 } 
   and 
     Modou Wade
     %\footnote{Supported by the MME-DII center of excellence (ANR-11-LABEX-0023-01)} 
 \end{center}

  \begin{center}
  { \it 
  Institut Camille Jordan, Université Jean Monnet, 23 Rue Dr Paul Michelon 42023 Saint-Étienne Cedex 2, France\\  
 THEMA, CY Cergy Paris Université, 33 Boulevard du Port, 95011 Cergy-Pontoise Cedex, France\\
  E-mail:   william.kengne@univ-st-etienne.fr  ; modou.wade@cyu.fr\\
  }
\end{center}

 \pagestyle{myheadings}
%\markboth{}{ Kengne and Wade}

\markboth{ Deep learning from strongly mixing observations: Sparse-penalized regularization and minimax optimality}{Kengne and Wade}

\medskip

\textbf{Abstract}:
The explicit regularization and optimality of deep neural networks estimators from independent data have made considerable progress recently.
The study of such properties on dependent data is still a challenge.
In this paper, we carry out deep learning from strongly mixing observations, and deal with a broad class of loss functions. 
We consider sparse-penalized regularization for deep neural network predictor. 
For a general framework that includes, regression estimation, classification, time series prediction$\cdots$ oracle inequalities for the expected excess risk are established, and upper bounds on the class of H\"older smooth functions and composition structured H\"older functions are provided.
For nonparametric autoregression with a Gaussian error and the Huber loss function, the upper bound of the $L_2$ error established matches (up to a logarithmic factor) with the lower bound established in \cite{kurisu2025adaptive} which shows that, the sparse-penalized deep neural networks estimator proposed is optimal in the minimax sense.
For autoregression with the Laplace error and the Huber loss function, an upper and a lower bound of the $L_2$ error on the class of composition H\"older functions are established.
These bounds match up to a logarithmic factor, that is, the proposed estimator attains the minimax optimal rate.
Based on the lower bound established in \cite{alquier2024minimax}, we show that the deep neural network estimator for the classification task with the logistic loss on strongly mixing observations is optimal in the minimax sense.
 
 \medskip
 

\medskip

\section{Introduction}
Recently, several research works have helped to understand the  theoretical properties of deep neural networks (DNN) estimators; see, for instance \cite{bauer2019deep}, \cite{schmidt2020nonparametric}, \cite{padilla2022quantile}, \cite{jiao2023deep} and the references therein.
In particular, the explicit regularization of DNN has received a considerable attention in the literature. 
Explicit regularization is based on modifying the objective of the optimization problem, unlike implicit regularization, which is induced by the learning algorithm.
The sparse-penalized and sparsity constraint approaches for independent and identical distribution (i.i.d.) data have been considered by several authors; see,  among other works, \cite{ohn2019smooth}, \cite{schmidt2020nonparametric}, \cite{tsuji2021estimation}, \cite{kim2021fast}, \cite{padilla2022quantile}, \cite{imaizumi2022advantage}, \cite{ohn2022nonconvex}, \cite{jiao2023deep}, \cite{shen2024nonparametric}. 
But, the i.i.d. assumption is quite restrictive and fails for many real life applications such as: market prediction, meteorological observations, medical bookings, signal processing$\dots$
We refer to \cite{kengne2024deep}, \cite{kurisu2025adaptive}, \cite{kengne2025excess}, \cite{chen2019bbs}, \cite{kohler2023rate},  \cite{ma2022theoretical}, \cite{kengne2024sparse}, \cite{men2024learning}, \cite{kengne2025robust} for some results obtained with dependent or non-i.i.d. observations. 

\medskip

 In this work, we consider the observations $ D_n \coloneqq  \{ (X_1, Y_1), \dots, (X_n, Y_n) \} $ (the training sample) which is a trajectory of a stationary and ergodic process $\{Z_t =(X_t, Y_t), t \in \Z \} $, which takes values in $\mathcal{Z} = \mathcal{X}  \times \mathcal{Y} \subset \R^d \times \R $ (with $d \in \N$), where $\mathcal{X}$ is the input space and $\mathcal{Y}$ the output space. 
 Let $\ell: \R \times  \mathcal{Y} \rightarrow [0, \infty)$ be a loss function that we use in the sequel. Set $\mathcal{F} \coloneqq \{h : \mathcal{X} \rightarrow \mathcal{Y}, ~ \text{measurable} \}$. For any predictor $ h\in \mathcal{F}$, we defined its risk by:
\begin{equation}\label{equa_def_risk}
 R(h) = \E_{Z_0} [\ell (h(X_0), Y_0)], ~ \text{with}~ Z_0 = (X_0, Y_0).  
\end{equation}
The target predictor (assumed to exist) $h^{*} \in \mathcal{F} $ fulfills:
\begin{equation}\label{best_pred_F}
R(h^{*}) =  \underset{h \in \mathcal{F}}{\inf}R(h).
\end{equation}
For any $h \in \mathcal{F}$, the excess risk is given by:
\begin{equation}\label{equa_excess_risk}
\mathcal{E}_{Z_0}(h) = R(h) - R(h^{*}), ~ \text{with} ~ Z_0 = (X_0, Y_0).
\end{equation} 

\noindent
We deal with a class of DNN estimators $\mathcal{H}_{\sigma}(L_n, N_n, B_n, F)$ (defined in Section \ref{Def_DNNs}, see (\ref{DNNs_no_Constraint})), with an activation function $\sigma$.  
The goal is to build from $D_n$, a DNN predictor $\widehat{h}_n \in \mathcal{H}_{\sigma}(L_n, N_n, B_n, F)$ with suitably chosen network architecture $(L_n, N_n, B_n, F)$, such that, its expected excess risk achieves the minimax optimal convergence rate.
We focus on the sparse-penalized regularization, and the  empirical minimizer over the class of DNN functions $\mathcal{H}_{\sigma}(L_n, N_n, B_n, F)$, also called sparse-penalized DNN (SPDNN) estimator is given by: 
\begin{equation}\label{sparse_DNNs_Estimators}
\widehat{h}_n = \underset{h \in  \mathcal{H}_{\sigma}(L_n, N_n, B_n, F)}{\argmin} \left[ \dfrac{1}{n} \sum_{i=1}^{n} \ell(h(X_i), Y_i) + J_n(h) \right],
\end{equation}
where $J_{n}(h) $ denotes the sparse penalty term, defined by: 
\begin{equation}\label{equa_penalty_term_v1}
J_n(h) = \sum_{j=1}^{p} \pi_{\lambda_n, \tau_n} \big( |\theta_j(h)| \big),
\end{equation}
where $\pi_{\lambda_n, \tau_n} : [0, \infty) \to [0, \infty)$ is a function with two tuning parameters $\lambda_n, \tau_n > 0$, and $\theta_j(h)$ is the $j$-th component of $\theta(h) = \big(\theta_1(h), \cdots, \theta_p(h) \big)^T$, which is the parameter of the network estimator $h \in \mathcal{H}_{\sigma}(L_n, N_n, B_n, F)$, and $^T$ denotes the transpose.
 We assume that $\pi_{\lambda_n, \tau_n}$ satisfies the two following conditions (see also \cite{kurisu2025adaptive}):
\begin{itemize}
\item [(i)] $\pi_{\lambda_n, \tau_n}(0) = 0$ and $\pi_{\lambda_n, \tau_n}(\cdot)$ is non-decreasing.
\item[(ii)] $\pi_{\lambda_n, \tau_n}(x) = \lambda_n$ if $x > \tau_n$.   
\end{itemize}
The clipped $L_1$ penalty of \cite{zhang2010analysis}, defined by for all $x \ge 0$ as:
\begin{equation}\label{equa_clipped_L1_penalty}
\pi_{\lambda_n, \tau_n}(x) = \lambda_n \Big(\dfrac{x}{\tau_n} \land 1 \Big), 
\end{equation}
considered also  by \cite{ohn2022nonconvex}, the SCAD penalty considered by \cite{fan2001variable}, the minimax concave penalty \cite{zhang2010nearly} or the seamless $L_0$ penalty in \cite{dicker2013variable}, are specific examples of the penalty function $J_n(h)$ in (\ref{equa_clipped_L1_penalty}) (see \cite{kurisu2025adaptive}).
%
%
%
%where $\theta= (\theta_1, \cdots, \theta_ p)^T $ is a  $p$-dimensional 
%vector and $^T$ denotes the transpose.
%
In the sequel, we set $\ell(h, z) = \ell(h(x), y)$ for all $z = (x, y) \in \mathcal{Z} = \mathcal{X}\times \mathcal{Y} $ and $h\in \mathcal{F}$.
We will be interested to find a bound of the expected excess risk and investigate if its convergence rate is minimax optimal.

\medskip

The minimax optimality of DNN estimators has already been considered in the literature. 
\cite{schmidt2020nonparametric} consider nonparametric regression with Gaussian error.
On a class of H\"older composition functions, they proved that, the DNN estimator with ReLU (rectified linear unit) activation function attains the minimax convergence rate, up to a logarithmic factor.   
Such results have been established by \cite{kurisu2025adaptive} for sparse-penalized DNN estimator for nonparametric autoregression with Gaussian error and by \cite{padilla2022quantile} for quantile regression with  ReLU DNN.
In this class of composition structured functions, sparse-penalized DNN predictor achieves the minimax optimal convergence rate for regression problem with sub-Gaussian error and classification, see \cite{ohn2022nonconvex}.  
We also refer to \cite{bauer2019deep}, \cite{hayakawa2020minimax}, \cite{imaizumi2019deep}, \cite{jiao2023deep}, \cite{shen2024nonparametric} and the references therein, for other results on the optimality of estimators based on DNN.      
 However, each of the aforementioned works is based on one or more of the following four assumptions: (i) the observations are independent; (ii) sparsity constrained regularization; (iii)  bounded loss function or output variable; (iv) nonparametric regression with Gaussian or sub-Gaussian error.  

 \medskip
 \noindent
 Regarding the sparsity constrained regularization, let us point out that, besides its discrete nature (leading to difficulties in the calculation of the estimator in practice), such estimator is not adaptive, since the sparsity level $S_n$ for attaining the minimax optimality rate depends on the smoothness of the target predictor $h^*$ which is unknown; see also the discussion in \cite{kurisu2025adaptive} (page 246) or \cite{ohn2022nonconvex} (pages 479-480).

 \medskip
 
 In this paper, we perform a sparse-penalized DNN predictor from a strongly mixing process $\{Z_t = (X_t, Y_t), t\in \Z\}$ which takes values in $\mathcal{Z} = \mathcal{X} \times \mathcal{Y} \subseteq \R^d\times \R$, based on the training sample $D_n = (X_1, Y_1), \dots, (X_n, Y_n)$ and a  loss function $\ell: \R\times \mathcal{Y} \rightarrow [0, \infty)$.
 Our main contributions focus on the following issues:
 
\medskip

\begin{enumerate}
\item \textbf{Oracle inequality and excess risk bounds of the SPDNN estimator}.
For a general framework, that includes regression and classification with a Lipschitz continuous loss function, oracle inequality for the expected excess risk of the SPDNN predictor is established. 
\begin{itemize}
\item \textbf{Excess risk bounds on the class of H\"older functions}.  
 The expected excess risk bound, when the target function belongs to the class of H\"older smooth functions (defined in Subsection \ref{excess_risk_Holder_space}, see (\ref{equa_ball_Holder})) with smoothness parameter $s$ is derived.
Under the subexponential strong mixing condition, the convergence rate of this bound is of order $\mathcal{O}\Big(  \big( n^{(\alpha)} \big)^{-\frac{\kappa s}{\kappa s + d} } \Big) \big(\log n^{(\alpha)}  \big)^{3} $ (where $n^{(\alpha)}$ is given in (\ref{n_alpha})) for some constant $\kappa \ge 1$, and $d$ is the dimension of the input space. 
Moreover, when the data are exponentially strongly mixing, the convergence rate of the bound of the expected excess risk is $\mathcal{O}\Big(n^{-\frac{\kappa s}{\kappa s + d} }  \big(\log (n)  \big)^{5} \Big)$.
For nonparametric regression with a symmetric (around 0) error and the Huber loss, and the classification problem with the logistic loss, these rates are obtained with $\kappa=2$.
\item \textbf{Excess risk bounds on the class of composition structured  H\"older functions}.
Under the subexponential strong mixing condition, we show that, when the target function belongs to the class of composition structured H\"older functions (defined in Subsection \ref{excess_risk_comp_Holder}, see (\ref{equa_compo_structured_function})), the convergence rate of the expected excess risk bound of the SPDNN estimator is of order $\mathcal{O} \Big( \big(\phi _{n^{(\alpha)}} ^{\kappa/2} \lor \phi _{n^{(\alpha)}}\big) \big(\log \big( n^{(\alpha)} \big)  \big)^{3} \Big)$, where $ \phi _{n^{(\alpha)}}$ is given in (\ref{def_phi_nalpha}). 
In this class, the convergence rate is $\mathcal{O}\Big( \big(\phi _{n} ^{\kappa/2} \lor \phi _{n}\big) \big(\log (n)  \big)^{5} \Big)$ when dealing with exponentially strongly mixing observations. 
As above, these rates can be applied with $\kappa = 2$ for nonparametric regression and classification.
\end{itemize}
\item \textbf{Minimax optimality in nonparametric autoregression and binary classification on the class of composition H\"older functions}. 
\begin{itemize}
    \item \textbf{Autoregression with a Gaussian error}. We deal with the Huber loss function. So, the upper bound of the expected $L_2$ error of the predictor is of order $\mathcal{O}\Big( \phi _{n} \big(\log (n)  \big)^{5} \Big)$, which matches (up to a logarithmic factor) with the lower bound established in \cite{kurisu2025adaptive} and shows that, the SPDNN estimator proposed is optimal in the minimax sense.
   \item \textbf{Autoregression with a Laplace error}. A lower bound of the $L_2$ error on the class of composition H\"older functions is established.
   For the learning task with the Huber loss, the upper bound of the expected $L_2$ error of the estimator matches (up to a logarithmic factor) with the lower bound obtained, which shows that, the SPDNN predictor is minimax optimal. 
   \item \textbf{Binary classification}. The problem of binary time series classification with the logistic loss is performed. 
   The upper bound of the expected excess risk of the predictor matches (up to a logarithmic factor) with the lower bound obtained in \cite{alquier2024minimax}, that is, the SPDNN estimator is optimal in the minimax sense.  
 \end{itemize} 
\end{enumerate}
\medskip

The rest of this paper is structured as follows. Some notations and assumptions are provided in Section \ref{asump}.  
Section \ref{Def_DNNs} presents the class of DNN considered. 
Section \ref{oracle_inqua_excess_risk} focuses on a general framework and establishes an oracle inequality and excess risk bound for the SPDNN predictor.
Minimax optimality in nonparametric autoregression and classification for strongly mixing observations are performed in Section \ref{nonparametric_regression}.
Section \ref{prove} is devoted to the proofs of the main results.

\medskip

\section{Notations and assumptions}\label{asump}
In the sequel, we set the following notations. Let $E_1$ and $E_2$ be two subsets of separable Banach spaces equipped respectively with a norm $\|\cdot\|_{E_1}, \|\cdot\|_{E_2}$.

\medskip

\noindent
For any function $h: E_1 \rightarrow E_2$ and $U \subseteq E_1$.
\begin{itemize}
\item We set,
\begin{equation}\label{def_norm_inf}
\| h\|_{\infty} = \sup_{x \in E_1} \| h(x) \|_{E_2}, ~ \| h\|_{\infty,U} = \sup_{x \in U} \| h(x) \|_{E_2}.
\end{equation}   
\item $\mathcal{F}(E_1, E_2)$ denotes the set of measurable functions from $E_1$ to $E_2$. 
\item For any $h\in \mathcal{F}(E_1, E_2), \epsilon >0$, $B(h, \epsilon) $ denotes the ball of radius $\epsilon$ of $\mathcal{F} (E_1, E_2) $ centered at $h$, that is,
\[ B (h, \epsilon) = \big\{ f \in \mathcal{F}(E_1, E_2), ~ \| f - h\|_\infty \leq \epsilon \big\},  \]
where $\| \cdot \|_\infty$ denotes the sup-norm defined in (\ref{def_norm_inf}).
\item  For any $\epsilon > 0$, the $\epsilon$-covering number $\mathcal{N}(\mathcal{H}, \epsilon)$  of $\mathcal{H} \subset \mathcal{F}(E_1, E_2) $, represents the minimal number of balls of radius $\epsilon$ needed to cover $\mathcal{H}$; that is,
\begin{equation}\label{epsi_covering_number}
 \mathcal{N}(\mathcal{H}, \epsilon) = \inf\Big\{ m \geq 1 ~: \exists h_1, \cdots, h_m \in \mathcal{H} ~ \text{such that} ~ \mathcal{H} \subset \bigcup_{i=1}^m B(h_i, \epsilon) \Big\}.
\end{equation}
\item For any $x \in \R$, $\lfloor x \rfloor$ denotes the greatest integer less than or equal to $x$, and $\lceil x \rceil$ the least integer greater than or equal to $x$. 
\item For all $u, v \in \R$, we set $u\lor v = \max(u, v)$ and $u \land v = \min(u, v)$.
\item For two sequences of real numbers $(u_n)$ and $(v_n)$, we write $ u_n \lesssim v_n$ or $ v_n \gtrsim u_n$ if there exists a constant $C > 0$ such that $u_n \leq  C v_n$ for all $n \in \N$; $u_n \asymp v_n$ if $u_n \lesssim v_n$ and $u_n \gtrsim v_n$.
\item For any finite set $A$, $|A|$ denotes the cardinal of $A$.
\item For a function $f$ defined on a set $E$, $f|_A$ denotes the restriction of $f$ to $A \subset E$. 
\end{itemize}

Let us recall the definition of strongly mixing processes, see \cite{rosenblatt1956central}, \cite{vidyasagar2013learning}.

\begin{Def}
Let $\mathcal{Z} = \{Z_i\}_{i\in \Z}$ be a stationary process on a probability space $(\Omega, \mathcal{B}, P)$. For $-\infty \leq i \leq \infty$, let $\sigma_{i}^{\infty}$ and $\sigma_{-\infty}^i$ be the sigma-algebras generated respectively by the random variables $Z_j, j \ge i$ and $Z_j, j\leq i$. The process $\mathcal{Z}$ is said to be $\alpha$-mixing, or strongly mixing, if
\begin{equation*}
\underset{A \in \sigma_{-\infty}^0, B \in\sigma_k^{\infty}}{sup} \{ |P(A \cap B) - P(A)P(B) |\} = \alpha(k) \to 0 \text{ when } k \to \infty. 
\end{equation*}
$\alpha(k)$ is then called the strong mixing coefficient.

\end{Def}

In the sequel, we consider the following definition, see also  \cite{ohn2019smooth}, \cite{ohn2022nonconvex}.
\begin{Def}\label{def_pwl_quad}
Let a function $g: \R \rightarrow \R$.
\begin{enumerate}
\item $g$ is continuous piecewise linear (or "piecewise linear" for notational simplicity) if it is continuous and there exists $K$ ($K\in \N$) break points $a_1,\cdots, a_K \in \R$ with $a_1 \leq a_2\leq\cdots \leq a_K $ such that, for any $k=1,\cdots,K$, $g'(a_k-) \neq g'(a_k+)$ and $g$ is linear on $(-\infty,a_1], [a_1,a_2],\cdots [a_K,\infty)$.
\item $g$ is locally quadratic if there exits an interval $(a,b)$ on which $g$ is three times continuously differentiable with bounded derivatives and there exists $t \in (a,b)$ such that $g'(t) \neq 0$ and $g''(t) \neq 0$.
\end{enumerate} 
\end{Def}

\medskip

We set the following assumptions on the process $\{ Z_t=(X_t, Y_t), t \in \Z \}$ which take values in $ \mathcal{Z}=   \mathcal{X} \times \mathcal{Y} \subset \R^d \times \R$, the input space $\mathcal{X}$, the loss function 
$\ell : \R \times \mathcal{Y} \to [0, \infty)$ and an activation function $\sigma: \R \rightarrow \R $, used in Section \ref{Def_DNNs}.
\begin{itemize}
\item[\textbf{(A0)}:] $\mathcal{X}\subset R^d$ is a compact set.
\item[\textbf{(A1)}:] There exists a constant $C_{\sigma}>0$ such that the activation function $\sigma$ is $C_{\sigma}$-Lipschitz, that is, there exists $C_{\sigma} > 0$ such that $|\sigma(x_1) - \sigma(x_2)| \leq C_{\sigma} |x_1 - x_2|$ for any $x_1, x_2 \in \R$.
Moreover, $\sigma$ is  either piecewise linear or locally quadratic and fixes a non empty interior segment $I \subseteq [0,1]$ (i.e. $\sigma(z) = z$ for all $z \in I$).
\item[\textbf{(A2)}:] There exists $\mk _{\ell} > 0$ such that, the loss function $\ell$ is $\mathcal{K}_{\ell}$-Lipschitz, that is, $\lip (\ell) \leq \mk _{\ell}$.
\item[\textbf{(A3)}:] The process $\{Z_t = (X_t, Y_t ),  t \in \Z \}$ is stationary and ergodic, and strong mixing  with the mixing coefficients defined by
\begin{equation}\label{coef_alpha_mixing}
\alpha(j) \leq \overline{\alpha} \exp(-c j^{\gamma}), ~ j \ge 1, \overline{\alpha} > 0, \gamma > 0, c > 0.
\end{equation}
\item[\textbf{(A4)}] 
Local structure of the excess risk: There exist three constants $\mk_0:= \mk_0(Z_0, \ell, h^*) , \varepsilon_0:= \varepsilon_0(Z_0, \ell, h^*) >0$ and $\kappa:=\kappa(Z_0, \ell, h^*) \geq 1$ such that,  
\begin{equation}\label{assump_local_quadr}
 R(h) - R(h^*) \leq \mk_0 \| h - h^*\|^\kappa_{\kappa, P_{X_0}}, 
\end{equation}
for any measurable function $h: \R^d \rightarrow \R$ satisfying $\|h - h^*\|_{\infty, \mx} \leq \varepsilon_0$; where $P_{X_0}$ denotes the distribution of $X_0$ and 
\[ \| h - h^{*}\|_{r, P_{X_0}} ^r := \displaystyle \int \| h (\text{x})- h^{*} (\text{x}) \|^r  d P_{X_0} ( \text{x}),  
\]
for all $r \geq 1$.
\end{itemize}

\medskip

\noindent
The assumptions above are satisfied in many learning frameworks:
\begin{itemize}
\item The ReLU (rectified linear unit) activation function defined by: $\text{ReLU}(x) = \max(x, 0)$ satisfies Assumption \textbf{(A1)} and also several other activation functions, see \cite{kengne2025excess}. 

\medskip

\noindent
The $L_1$ and the Huber loss with parameter $\delta >0$ satisfy Assumption \textbf{(A2)} with $\mathcal{K}_{\ell} = 1$ and $\mathcal{K}_{\ell} = \delta$ respectively. 
For all $y \in \R$ and $y' \in \mathcal{Y}$, recall: 
\item The $L_1$ loss function: $\ell(y,y') = |y-y'|$;
\item The Huber loss with parameter $\delta >0$:
  \begin{equation}\label{def_Huber_loss_v1}
\ell(y,y') =  \left\{
\begin{array}{ll}
      \frac{1}{2}(y-y')^2 &  \text{ if } ~ |y-y'|\leq \delta , \\
 \delta |y-y'| - \frac{1}{2} \delta^2 &   ~ \text{ otherwise}.
\end{array}
\right.      
\end{equation} 
\item Several classical autoregressive processes satisfy Assumption \textbf{(A3)}, see for example \cite{doukhan1994mixing}, \cite{chen2000geometric} and Section \ref{nonparametric_regression} below. 
\item Assumption \textbf{(A4)} is satisfied with $\kappa =1$
when the loss function is Lipschitz continuous. 
One can easily see from Lemma C.6 in \cite{zhang2024classification} that Assumption \textbf{(A4)} is satisfied with $\kappa = 2$ for binary classification using the logistic loss.
In the case of regression with the Huber loss, if the error is symmetric around 0, Assumption \textbf{(A4)} holds with $\kappa = 2$; see Proposition 3.1 in \cite{fan2024noise}.
\end{itemize}

\medskip

\section{ Deep Neural Networks }\label{Def_DNNs}
A DNN function with network architecture $ (L, \textbf{p}) $, where $L$ stands for the number of hidden layers or depth and $\mathbf{p} = (p_0, p_1,\cdots, p_{L+1}) \in \N^{L+2}$ denotes a width vector can be expressed as a function of the form:
\begin{equation}\label{h_equ1}
h: \R^{p_0} \rightarrow \R^{p_{L+1}}, \;   x\mapsto h(x) = A_{L+1} \circ \sigma_{L} \circ A_{L} \circ \sigma_{L-1} \circ \cdots \circ \sigma_1 \circ A_1 (x),
\end{equation} 
where $ A_j: \R^ {p_ {j -1}} \rightarrow \R^ {p_j} $ is a linear affine map, defined by $A_j (x) \coloneqq W_j x + \textbf{b}_j$,  for given  $p_ {j}\times p_{j-1}$  weight matrix   $ W_j$   and a shift vector $ \textbf{b}_j \in \R^ {p_j} $, and $\sigma_j: \R^{p_j} \rightarrow \R^ {p_j} $ is a nonlinear element-wise activation map, defined for all $z=(z_1,\cdots,z_{p_j})^T$ by $\sigma_j (z) = (\sigma(z_1), \cdots, \sigma(z_{p_j}))^{T} $, and $^T$ denotes the transpose.   
Let us denote by,
\begin{equation} \label{def_theta_h}
\theta(h) \coloneqq \left(vec(W_1)^ {T}, \textbf{b}^{T}_{1}, \cdots,  vec(W_{L + 1})^{T} , \textbf{b}^ {T}_{L+1} \right)^{T}, 
 \end{equation} 
 the vector of parameters for a DNN function of the form  (\ref{h_equ1}), where $ vec(W)$ is the vector obtained by concatenating the column vectors of the matrix $W$. 
 In the sequel, we consider an activation function $ \sigma: \R \rightarrow \R$ and let us denote by $\mathcal{H}_{\sigma, p_0, p_{L+1}} $, the class of DNN predictors with $p_0$-dimensional input and $p_{L+1} $-dimensional output.
Set $p_0 = d$ and $p_ {L + 1} = 1$ in the setting considered here.
For a DNN $h$ as in (\ref{h_equ1}), denote by depth($h$) and width($h$) its depth and width respectively; that is, depth($h$)$=L$ and width($h$) = $\underset{1\leq j \leq L} {\max} p_j $. For any positive constants $L, N, B, F > 0$, set
\[ \mathcal{H}_{\sigma}(L, N, B) \coloneqq \big \{h\in \mathcal{H}_{\sigma, d, 1}: \text{depth}(h)\leq L, \text{width}(h)\leq N, \|\theta(h)\|_{\infty} \leq B \big\},  \]
 and
\begin{equation}\label{DNNs_no_Constraint}
\mathcal{H}_{\sigma}(L, N, B, F) \coloneqq \big\{ h: h\in H_{\sigma}(L, N, B), \| h \|_{\infty, \mathcal{X}} \leq F \big\}. 
\end{equation}
The class of sparsity constrained DNNs with sparsity level $S > 0$, is defined by: 
\begin{equation}\label{DNNs_Constraint}
\mathcal{H}_{\sigma}(L, N, B, F, S) \coloneqq \left\{h\in \mathcal{H}_{\sigma}(L, N, B, F) \; :  \; \| \theta(h) \|_0 \leq S 
  \right\},
\end{equation}
where $\| x \|_0 = \sum_{i=1}^p \ind(x_i \neq 0), ~ \| x\|_{\infty} = \underset{1 \leq i  \leq p}{\max} |x_i |$ for all $x=(x_1,\ldots,x_p)^{T} \in \R^p$ ($p \in \N$).
In the sequel, we deal with network architecture parameters $L, N, B, F, S$ that depend on the sample size $n$.

\section{Oracle inequality and excess risk bounds in a general framework}\label{oracle_inqua_excess_risk}
In this section, we establish an oracle inequality and derive bounds of the expected excess risk for the SPDNN estimator under a strong mixing condition. 
This involves a calibration of the network architecture parameters $L_n, N_n, B_n, F$ and the tuning parameters $\lambda_n, \tau_n$ of the penalty term.

\subsection{Oracle inequalities for the excess risk}
 
The following theorem provides an oracle inequality for the expected excess risk of the SPDNN estimator from subexponential strongly mixing observations.

\begin{thm}\label{thm1}
Assume that (\textbf{A1})-(\textbf{A3}) hold.  Let $ L_n \asymp\log\big(n^{(\alpha)}\big), N_n \lesssim \big(n^{(\alpha)}\big)^{\nu_1}, 1 \leq B_n \lesssim \big(n^{(\alpha)}\big)^{\nu_2}, F_n = F > 0$, for some $ \nu_1 > 0, \nu_2 > 0$.  Let $\nu_3>2$ and $h^{*}$ defined in (\ref{best_pred_F}) is such that $\|h^{*}\|_{\infty} \leq \mathcal{K}$ for some constant $\mathcal{K} > 0$. Then, there exists $n_0 \in \N$ such that, for all $n \geq n_0$, the SPDNN estimator defined by,
\begin{equation}\label{EMRpen_v1}
\widehat{h}_n = \underset{h\in \mathcal{H}_{\sigma} (L_n, N_n, B_n, F)}{\argmin}\Big[\dfrac{1}{n} \sum_{i = 1}^{n} \ell \left( h(X_i), Y_i \right) + J_n(h)\Big],
\end{equation}
with $\lambda_n \asymp \frac{\big( \log(n^{(\alpha)}) \big) ^{\nu_3}}{n^{(\alpha)}}, \tau_n \leq \dfrac{1}{16 \mathcal{K}_{\ell}(L_n + 1)((N_n + 1)B_n)^{L_n + 1} n^{(\alpha)} }$ satisfies,
\begin{align}
\nonumber \E\big[\mathcal{E}_{Z_0}(\widehat{h}_n) \big]
&  \leq 2\Big(\underset{h \in \mathcal{H}_{\sigma}(L_n, N_n, B_n, F)} {\inf} \Big[ \mathcal{E}_{Z_0} (h)  + J_n(h) \Big]  \Big)  + \dfrac{\Xi}{n^{(\alpha)}},
\end{align}
for some constant $\Xi>0$, where $ \mathcal{K}_{\ell} > 0$ is given in Assumption (\textbf{A2}) and $n^{(\alpha)}$ defined by
\begin{equation}\label{n_alpha}
n^{(\alpha)} \coloneqq \lfloor n \lceil \{8n/c\}^{1/(\gamma + 1)}\rceil^{-1} \rfloor,
\end{equation}
for some positive constant $c > 0, \gamma > 0$ defined in (\ref{coef_alpha_mixing}).
\end{thm}

\medskip

The next theorem establishes an oracle inequality for the expected excess risk of the SPDNN estimator under exponential strong mixing condition.
\begin{thm}\label{oracle_inequality_expo_strong_mixing}
Assume that \textbf{(A1 )}, \textbf{(A2)} and \textbf{(A3)} with $\gamma \ge 1$ hold. Let $ L_n \asymp\log(n), N_n \lesssim n^{\nu_1}, 1 \leq B_n \lesssim n^{\nu_2}, F_n = F > 0$, for some $ \nu_1 > 0, \nu_2 > 0$.  Let $\nu_3 > 4$ and $h^{*}$ defined in (\ref{best_pred_F}) is such that $\|h^{*}\|_{\infty} \leq \mathcal{K}$ for some constant $\mathcal{K} > 0$. Then, there exists $n_0 \in \N$ such that, for all $n \geq n_0$, the SPDNN estimator defined in (\ref{EMRpen_v1}),
with $\lambda_n \asymp \frac{\big(\log n
 \big) ^{\nu_3}}{n}, \tau_n \leq \dfrac{1}{16 \mathcal{K}_{\ell}(L_n + 1)((N_n + 1)B_n)^{L_n + 1} n}$ satisfies,
\begin{align}
\nonumber \E\big[\mathcal{E}_{Z_0}(\widehat{h}_n) \big]
&  \leq 2\Big(\underset{h \in \mathcal{H}_{\sigma}(L_n, N_n, B_n, F)} {\inf} \Big[ \mathcal{E}_{Z_0} (h)  + J_n(h) \Big]  \Big)  + \dfrac{\Xi (\log n) ^2}{n},
\end{align}
for some constant $\Xi>0$, where $ \mathcal{K}_{\ell} > 0$ is given in Assumption (\textbf{A2}). 
\end{thm}

\noindent
The inequalities above can be used to establish bounds of the expected excess risk of the SPDNN predictor.

\subsection{Excess risk bounds on the class of H\"older functions}\label{excess_risk_Holder_space}
 Recall that for any $r\in \N, D\subset \R^r, \beta, A >0$, the ball of $\beta$-H\"older functions with radius $A$ is defined by:
\begin{equation}\label{equa_ball_Holder}
 \mathcal{C}^{\beta}(D, A)= \Bigg\{h : D \rightarrow \R : \underset{\boldsymbol{\alpha} : |\boldsymbol{\alpha}|_1 < \beta}{\sum}\|\partial^{\boldsymbol{\alpha}}h\|_{\infty} + \underset{\boldsymbol{\alpha} : |\boldsymbol{\alpha}|_1 = \lfloor \beta \rfloor}{\sum}~\underset{x\ne y}{\underset{x, y\in D}{\sup}} \frac{|\partial^{\boldsymbol{\alpha}}h(x) - \partial^{\boldsymbol{\alpha}}h(y)|}{|x - y|^{\beta - \lfloor \beta \rfloor}} \leq A \Bigg\},  
\end{equation}
with $\boldsymbol{\alpha} = (\alpha_1, \dots, \alpha_r)\in \N^r$, $|\boldsymbol{\alpha}|_1 \coloneqq \sum_{i=1}^r \alpha_i$ and $\partial^{\boldsymbol{\alpha}} = \partial^{\alpha_1} \dots \partial^{\alpha_r}$.

\medskip

\noindent
The following corollary provides an upper bound for the excess risk of the SPDNN estimator under subexponential and exponential strong mixing conditions.

\begin{Corol}\label{corol1}
Assume that \textbf{(A0)}, \textbf{(A1)}, \textbf{(A2)}, \textbf{(A4)} for some $\kappa \ge 1, \mk_0, \varepsilon_0 > 0$ hold and that 
 $h ^* \in \mathcal{C}^s(\mathcal{X}, \mathcal{K})$ for some $s, \mathcal{K} > 0$ the class of H\"older smooth functions given in (see (\ref{equa_ball_Holder})).
 Let us consider the DNNs class $\mathcal{H}_{\sigma}(L_n, N_n, B_n, F)$ with network architecture $(L_n, N_n, B_n, F)$ and an activation function $\sigma$.
 \begin{enumerate}
\item Assume that \textbf{(A3)} fulfills. Let $L_n \asymp \log\big(n^{(\alpha)}\big), N_n \asymp \big(n^{(\alpha)}\big)^{\frac{d}{\kappa s + d}}, B_n \asymp \big(n^{(\alpha)}\big)^{\frac{4(s + d)}{\kappa s + d}}, F_n = F >0$ and $\lambda_n, \tau_n$ given as in Theorem (\ref{thm1}).
Then, the SPDNN estimator defined in (\ref{EMRpen_v1}) satisfies,
\begin{equation}\label{excess_risk_bound_v1}
\underset{h^{*} \in  \mathcal{C}^{s}(\mathcal{X}, \mathcal{K})}{\sup}\E[\mathcal{E}_{Z_0}(\widehat{h}_n) ] \lesssim \dfrac{  \big(\log\big(n^{(\alpha)}\big) \big)^{\nu} }{\big(n^{(\alpha)}\big)^{\kappa s/(\kappa s + d)}},
\end{equation}
for all $\nu > 3$ and $n^{(\alpha)}$ defined in (\ref{n_alpha}).
\item Assume that \textbf{(A3)} hold with $\gamma \ge 1$. Let $L_n \asymp \log(n), N_n \asymp n ^{\frac{d}{\kappa s + d}}, B_n \asymp n^{\frac{4(s + d)}{\kappa s + d}}, F_n = F >0$ and $\lambda_n, \tau_n$ given as in Theorem (\ref{oracle_inequality_expo_strong_mixing}).
Then, the SPDNN estimator defined in (\ref{EMRpen_v1}) satisfies,
\begin{equation}\label{excess_risk_bound_v2}
\underset{h^{*} \in  \mathcal{C}^{s}(\mathcal{X}, \mathcal{K})}{\sup}\E[\mathcal{E}_{Z_0}(\widehat{h}_n) ] \lesssim \dfrac{  \big(\log n\big) ^{\nu} }{n ^{\kappa s/(\kappa s + d)}},
\end{equation}
for all $\nu > 5$. 
\end{enumerate}
\end{Corol}

\medskip

\noindent 
\begin{rmrk} \label{rem1_bound} \hfill
\begin{itemize}
\item The bound (\ref{excess_risk_bound_v1}) shows that the convergence rate of the expected excess risk of the SPDNN predictor under subexponential strong mixing data is of order $\mathcal{O}\Big(  \big(n^{(\alpha)} \big)^{-\frac{\kappa s}{\kappa s + d}} \big( \log \big(n^{(\alpha)}\big) \big) ^3  \Big)$. 

\item For $n \ge \max(c/8, 2^{2 + 5/\gamma}c^{1/\gamma})$, we have $n^{(\alpha)} \ge C(\gamma, c) n^{\gamma/(\gamma+1)}$, with $C(\gamma,c) = 2^{-\frac{2\gamma + 5}{\gamma + 1}}c^{\frac{1}{\gamma + 1}}$, see also \cite{hang2014fast}.
So, the convergence rate in (\ref{excess_risk_bound_v1}) is of order $\mathcal{O}\Big(n^{-\big(\frac{\gamma}{\gamma + 1}\big)\big(\frac{\kappa s}{\kappa s + d}\big)}  \big(\log n \big)^3 \Big)$.
Hence, for i.i.d data, (that is $\gamma = \infty$), this rate is equal to  $\mathcal{O}\Big(n^{-\big(\frac{\kappa s}{\kappa s + d}\big)}  \big(\log n \big)^3 \Big)$.
\item  The bound (\ref{excess_risk_bound_v2}) shows that the convergence rate of the expected excess risk of the SPDNN predictor under exponential strong mixing data is of order $\mathcal{O}\Big(  n^{-\frac{\kappa s}{\kappa s + d}} (\log n) ^5 \Big)$.  
\end{itemize}
\end{rmrk} 

\subsection{Excess risk bounds on the class of composition  H\"older functions}\label{excess_risk_comp_Holder}
\textbf{Composition structured H\"older functions}. Let $q\in \N, \boldsymbol{d} = (d_0, \dots, d_{q+1})\in \N^{q + 2}$ with $d_0 = d, d_{q+1} = 1, \boldsymbol{t} = (t_0, \dots, t_q)\in \N^{q+1}$ with $t_i \leq d_i$ for all $i$, $\boldsymbol{\beta} = (\beta_0, \dots, \beta_q)\in (0,\infty)^{q+1}$ and $A>0$. 
For all $l < u$, denote by $\mathcal{C}_{t_i}^{\beta_i}([l, u]^{d_i}, A)$ the set of functions $f: [l, u]^{d_i} \rightarrow \R$ that depend on at most $t_i$ coordinates and $f \in \mathcal{C}^{\beta_i}([l, u]^{d_i}, A)$.

\medskip

\noindent Define, see \cite{schmidt2020nonparametric}, the set of composition structured functions $\mathcal{G}(q, \bold{d}, \bold{t}, \boldsymbol{\beta}, A)$ by: 
\begin{multline}\label{equa_compo_structured_function}
\mathcal{G}(q, \bold{d}, \bold{t}, \boldsymbol{\beta}, A) \coloneqq \Big\{h = g_q\circ \dots \circ g_0, g_i = (g_{ij})_{j=1,\cdots,d_{i+1}} : [l_i, u_i]^{d_i}
 \rightarrow [l_{i+1}, u_{i+1}]^{d_{i+1}}, g_{ij}\in C_{t_i}^{\beta_i}
 ([l_i, u_i]^{d_i}, A) ~ \\
        \text{for some } l_i, u_i \in \R \text{ such that } ~ |l_i|, |u_i| \leq A, \text{ for } i=1,\cdots,q \Big\}.
\end{multline}
The smoothness of a composition function in $\mathcal{G}(q, \bold{d}, \bold{t}, \boldsymbol{\beta}, A)$ (see (\cite{juditsky2009nonparametric})) are defined by
$\beta_i^* \coloneqq \beta_i\prod_{j = i+1}^q(\beta_{j} \land 1)$. We defined in the sequel by
\begin{equation}\label{def_phi_nalpha}
\phi_{n^{(\alpha)}} \coloneqq  \underset{0\leq i \leq q}{\max}\big(n^{(\alpha)} \big)^{ \frac{-2\beta_i^* }{2\beta_i^* + t_i}   }.
\end{equation}

\medskip

\noindent
The next corollary is established with the ReLU activation function, that is $\sigma(x) = \max(x, 0)$ (that satisfies \textbf{(A1)} with $C_\sigma =1$) and $\mathcal{X} = [0,1]^d$.
\begin{Corol}\label{corol01_v1}
Set $\mathcal{X} = [0,1]^d$. Assume that (\textbf{A2}),  (\textbf{A4}) for some $\kappa \ge 1, \mk_0 , \varepsilon_0 > 0$ hold and that $h ^* \in \mathcal{G}(q, \bold{d}, \bold{t}, \boldsymbol{\beta}, \mathcal{K})$ a class of composition structured functions defined in (\ref{equa_compo_structured_function}). Consider the DNNs class $\mathcal{H}_{\sigma}(L_n, N_n, B_n, F)$ with network architecture $(L_n, N_n, B_n, F)$.
\begin{enumerate}
\item Assume that (\textbf{A3}) is satisfied. Let $L_n\asymp \log\big(n^{(\alpha)}\big), N_n \asymp n^{(\alpha)} \phi_{n^{(\alpha)}}, B_n=B \geq 1 , F_n = F > \max(\mathcal{K}, 1)$ and $\lambda_n, \tau_n$ be chosen as in Theorem (\ref{thm1}).
Then, the SPDNN estimator defined in (\ref{EMRpen_v1}) with the ReLU activation function, satisfies,  
\begin{align}\label{excess_risk_bound_corol01_v1}
\underset{h^{*} \in \mathcal{G}(q, \bold{d}, \bold{t}, \boldsymbol{\beta}, \mathcal{K})}{\sup} \E[\mathcal{E}_{Z_0}(\widehat{h}_n) ]  & \lesssim \Big( \phi_{n^{(\alpha)}} ^{\kappa/2} \lor \phi_{n^{(\alpha)}} \Big) \big(\log\big(n^{(\alpha)} \big) \big)^{\nu},  
\end{align}
for all $\nu > 3$, where $\phi_{n^{(\alpha)}}$ is given in (see (\ref{def_phi_nalpha})).
\item Assume that (\textbf{A3}) with $\gamma \ge 1$ hold. Let $L_n\asymp \log (n), N_n \asymp n \phi_n, B_n=B \geq 1 , F_n = F > \max(\mathcal{K}, 1)$ and $\lambda_n, \tau_n$ be chosen as in Theorem (\ref{oracle_inequality_expo_strong_mixing}).
Then, the SPDNN estimator defined in (\ref{EMRpen_v1}) with the ReLU activation function, satisfies,  
\begin{align}\label{excess_risk_bound_corol01_v1}
\underset{h^{*} \in \mathcal{G}(q, \bold{d}, \bold{t}, \boldsymbol{\beta}, \mathcal{K})}{\sup} \E[\mathcal{E}_{Z_0}(\widehat{h}_n) ]  & \lesssim  \Big( \phi_{n} ^{\kappa/2} \lor \phi_{n} \Big) \big(\log(n) \big)^{\nu},  
\end{align}
for all $\nu > 5$.
\end{enumerate}
\end{Corol}
\noindent
When $\kappa=2$ in (\ref{excess_risk_bound_corol01_v1}), this rate coincides (up to a logarithmic factor) with that obtained in the i.i.d. cases, see for instance \cite{schmidt2020nonparametric}, \cite{padilla2022quantile}, \cite{ohn2022nonconvex}.

\medskip

\section{Minimax optimality for nonparametric autoregression and classification}\label{nonparametric_regression}
\subsection{Nonparametric autoregression}
In this subsection, we perform the SPDNN estimator for nonparametric autoregression. We consider nonlinear autoregressive models, where the process $(Y_{t})_{t\in \Z}$ which takes values in $\R$ satisfies:
\begin{equation}\label{equa_model_auto}
 Y_t = h^{*}(Y_{t-1}, \dots, Y_{t - d}) + \xi_t,    
\end{equation}
for some measurable function $h^{*} : \R^d \rightarrow \R$ ($d\in \N$), $(\xi_t)_{t\in \Z}$ is an i.i.d. centered random process, generated from either a standard Laplace or a standard Gaussian distribution. Set $X_t = (Y_{t-1}, \dots, Y_{t - d})$.
In this subsection, we set $\mathcal{X}=[0,1]^d$ and perform nonparametric estimation of $h^{*}$ on $[0,1]^d$; that is, an estimation of $h_{0}^{*} \coloneqq h^{*} \ind_{[0,1]^d}$. Let us check the assumption (\textbf{A3}) for the model (\ref{equa_model_auto}). To do this, we set the following assumptions.
\begin{itemize}
\item[A$_1(h^{*})$:] The function $h^{*}$ is bounded on every compact of $\R^d$, that is, for every $K \ge 0$, 
\begin{equation}\label{equa_cond_bouned}
 \underset{\|x\| \leq K}{\sup} |h^{*}(x)| < \infty.    
\end{equation}
\item[A$_2(h^{*})$:] There exist constants $\theta_i \ge 0, ~ i = 1, \cdots, d, c_0 > 0, \Gamma> 0$ such that 
\begin{equation}\label{equa_cond_lipschitz}
|h^{*}(x)| \leq \sum_{i = 1}^d \theta_i |x_i| + c_0, ~ \|x\| \ge \Gamma,
\end{equation}
\end{itemize}
and that 
\begin{equation}\label{equa_cond_const}
\sum_{i = 1}^d \theta_i < 1.
\end{equation}
In the sequel, we set:
\begin{equation}\label{def_cal_M}
\mathcal{M} := \big\{ h: \R^d \rightarrow \R,\text{ satisfying }  A_1(h) \text{ and } A_2(h) \text{ with }  (\ref{equa_cond_const}) \big\}.
\end{equation}
Under the assumptions A$_1(h^{*})$, A$_2(h^{*})$ and the condition (\ref{equa_cond_const}), there exists a stationary solution $(Y_t)_{t\in \Z}$ of the model (\ref{equa_model_auto}) which is geometrically strong mixing, see, for instance, \cite{chen2000geometric}, \cite{doukhan1994mixing}. Thus, the assumption (\textbf{A3}) is satisfied. In addition:
\begin{itemize}
\item[$\bullet$] We deal with the ReLU activation function, thus, the assumption (\textbf{A1}) holds.
\item[$\bullet$] We use the Huber loss function, thus, the assumption (\textbf{A2}) is satisfied.
\item[$\bullet$] We consider the assumption \textbf{(A3)} with $\gamma \ge 1$.
\end{itemize}
For the model (\ref{equa_model_auto}), with the Gaussian or standard Laplace error (which are symmetric around $0$), assumption \textbf{(A4)} holds with $\kappa = 2$.
Hence, the results of the second attempt of Corollary \ref{corol01_v1} can be applied to the model (\ref{equa_model_auto}).

\medskip

Let $\mathcal{G}(q, \bold{d}, \bold{t}, \boldsymbol{\beta}, \mathcal{K})$ be a class of composition structured functions defined in (\ref{equa_compo_structured_function}) for some $\mathcal{K} >0$.
Consider model (\ref{equa_model_auto}) and assume that A$_1(h^{*})$, A$_2(h^{*})$ hold and that $h_0^{*} \in \mathcal{G}(q, \bold{d}, \bold{t}, \boldsymbol{\beta}, \mathcal{K})$.
Let $\mathcal{H}_{\sigma}(L_n, N_n, B_n, F_n)$ be a class DNN with networks architecture parameters $L_n, N_n, B_n, F_n=F$ given as in the second attempt of Corollary \ref{corol01_v1} and where $\sigma$ is the ReLU activation function.
Furthermore, since the distribution of $\xi_0$ is symmetric around $0$, from Proposition $3.1$ in \cite{fan2024noise}, one can find a constant $C \ge 4$ such that, for the Huber loss function with parameter $\delta \ge C$, we have,
\begin{equation}
    \|h - h_0 ^*\|_{2, P_{X_0}} ^2 \leq 4 \big(R _{\delta}(h) -  R _{\delta} ( h_0 ^*) \big),
\end{equation}
for all bounded function $h$ on $\mx$, where $R _{\delta}$ is the risk with the Huber loss with parameter $\delta$.
Then, from the second attempt of Corollary \ref{corol01_v1}, the SPDNN estimator defined by
\begin{equation}\label{equa_SPDNN}
\widehat{h}_n = \underset{h\in \mathcal{H}_{\sigma}(L_n, N_n, B_n, F)}\argmin\Big[\frac{1}{n}\sum_{i=1}^{n} \ell_{\delta} \big(h(X_i), Y_i \big) + J_n(h)\Big],    
\end{equation}
where $\ell_{\delta}(\cdot,\cdot)$ stands for the Huber loss function with parameter $\delta > 0$ defined in (\ref{def_Huber_loss_v1}),
satisfies:
\begin{equation}\label{equa_l1_excess_risk}
\underset{h_0^{*} \in \mathcal{G}(q, \bold{d}, \bold{t}, \boldsymbol{\beta}, \mathcal{K})}{\sup}\E\big[\|\widehat{h}_n - h_0^{*}\|_{2, P_{X_0}}^2 \big]  \lesssim \phi_n \big(\log(n) \big)^{\nu} ,  
\end{equation}
for all $\nu > 5$, where $\phi_n$ is defined in (\ref{def_phi_nalpha}), and $h_{0}^{*} \coloneqq h^{*} \ind_{[0,1]^d}$.
Thus, the convergence rate in (\ref{equa_l1_excess_risk}) matches (up to a logarithmic factor) with the lower bound established in \cite{kurisu2025adaptive}. Which shows that, the SPDNN estimator $\widehat{h}_n$ with the Huber loss, for the nonparametric autoregression with the Gaussian error, is optimal in the minimax sense.

\medskip

Now, let us focus to the model (\ref{equa_model_auto}) with the standard Laplace error.
For this purpose, we first establish the following lemma:

\begin{lem}\label{equa_lem1}
Consider the class of composition structured functions $\mathcal{G}(q, \bold{d}, \bold{t}, \boldsymbol{\beta}, \mathcal{K})$ for sufficiently large $\mk >0$, and such that $t_i \leq \min(d_0, \dots, d_{i-1})$ for all $i$.
Then, for some integer $M \geq 1$, there exists $M+1$ functions, $h_{(0)}, \dots, h_{(M)} \in \mathcal{G}(q, \bold{d}, \bold{t}, \boldsymbol{\beta}, \mathcal{K})$ satisfying for any $n\geq 1$:
\begin{itemize}
\item[(i)] $\|h_{(j)} - h_{(k)}\|_{L^2[0,1]^d} \ge \kappa \sqrt{\phi_{n}}$ for all $0 \leq j < k \leq M$;
\item[(ii)] $\frac{1}{M}\sum_{j=1}^M n\|h_{(j)} - h_{(0)}\|_{L^2[0,1]^d} \leq \log(M)/9$,
\end{itemize}
for some constant $\kappa >0$, independent of $n$ and with,
\begin{equation}\label{def_phi_n}
\phi_{n} \coloneqq  \underset{0\leq i \leq q}{\max} n^{ \frac{-2\beta_i^* }{2\beta_i^* + t_i}   }.
\end{equation}
 
\end{lem}

\medskip

For a class composition functions $\mathcal{G}(q, \bold{d}, \bold{t}, \boldsymbol{\beta}, A)$ defined in (\ref{equa_compo_structured_function}), set:
\begin{equation}
 \mathcal{M}(q, \bold{d}, \bold{t}, \boldsymbol{\beta}, A) := \big\{ h \in \mathcal{M}, ~ \text{ such that } h\ind_{[0,1]^d} \in   \mathcal{G}(q, \bold{d}, \bold{t}, \boldsymbol{\beta}, A) \big\},
\end{equation} 
where $\mathcal{M}$ is defined in (\ref{def_cal_M}).
The following Theorem establishes a lower bound of the expected $L_2$ error.
\begin{thm}\label{thm2}
Consider the nonparametric autoregression model (\ref{equa_model_auto}) with i.i.d. standard Laplace error process $(\xi_t)$.
Consider a class of composition structured functions $\mathcal{G}(q, \bold{d}, \bold{t}, \boldsymbol{\beta}, \mathcal{K})$ with $t_i \leq \min(d_0, \dots, d_{i-1})$ for all $i$ and $\mathcal{M}(q, \bold{d}, \bold{t}, \boldsymbol{\beta}, \mathcal{K})$.
 Then for sufficiently large $\mk$, there exist a constant $C>0$  such that 
\begin{equation}\label{equa_lower_bound}
\underset{\widehat{h}_n}{\inf}~\underset{h_0^{*}\in\mathcal{M}(q, \bold{d}, \bold{t}, \boldsymbol{\beta}, \mk)}{\sup} \E[\|\widehat{h}_n - h_0^{*}\|_{2, P_{X_0}}^2] \ge C\phi_n,
\end{equation}
where the $\inf$ is taken over all estimator based on $(Y_1,\cdots, Y_n)$ generated from (\ref{equa_model_auto}) and $\phi_n$ is defined in (\ref{def_phi_nalpha}).
\end{thm}

\medskip

\noindent
Thus, the convergence rate of the lower bound of the $L_2$ error in (\ref{equa_lower_bound}) matches, up to a logarithmic factor, the convergence rate of the upper bounds (\ref{equa_l1_excess_risk}). 
Thus, the SPDNN estimator for nonparametric autoregression model with standard Laplace error with target function in the class $\mathcal{M}(q, \bold{d}, \bold{t}, \boldsymbol{\beta}, \mk)$, is optimal (up to a logarithmic factor) in the minimax sense.

\medskip

\noindent
Let us consider one example of the model below.
\begin{example}[Generalized exponential autoregressive models] Consider the nonlinear autoregressive model: 
\begin{equation}\label{equa_GEXPAR}
\textbf{GEXPAR}: ~ Y_t =h^*(Y_{t-1}, \cdots, Y_{t-d}) + \xi_t= c_0 + \sum_{i=1}^d\big(c_i + \pi_i e^{\lambda(Y_{t - i} - z_i)^2}\big)Y_{t - i} + \xi_t, ~ \forall t\in \Z, 
\end{equation}
for parameters $z_i,c_i, \pi_i, \lambda \in \R$,  $i=1, \dots, d$, $d\in \N$ and $(\xi_t)_{t\in \Z}$ is a sequence of i.i.d. centered random variables generated from either standardized Laplace or normal distribution. One can see that the model (\ref{equa_GEXPAR}) is a particular case of the model (\ref{equa_model_auto}). Set $X_t = (Y_{t-1}, \dots, Y_{t-d})$. 

Let us check stability conditions of the \textbf{GEXPAR} models. To do this, we set the following assumption:
\begin{itemize}
%\item A($\xi_t$): The sequence $(\xi_t)_{t\in \Z}$ has a positive density everywhere on the real line $\R$.
%
\item[] A($\varphi$):  The characteristic equation (\ref{equa_characteristic}) defined below have all roots inside the unit circle 
\begin{equation}\label{equa_characteristic}
z^d - \varphi_1 z^{d-1} - \dots - \varphi_{d-1}z - \varphi_d = 0,
\end{equation}
where $\varphi_i = |c_i| + |\pi_i|$ for $i=1, \dots, d$.
\end{itemize}
Under the assumption  A($\varphi$), there exists a stationary solution of the model (\ref{equa_GEXPAR}) which is geometrically strongly mixing, see Theorem $2.1$ in \cite{chen2018generalized} and Theorem $1$ in \cite{chen2000geometric}.   

\medskip

\noindent
Recall that we consider the Huber loss function, and the target function with respect to this loss function in the case of model (\ref{equa_GEXPAR}) (with symmetric error around 0) is $h^*(x), ~ for ~ x\in \R^d$. 
Thus, the assumptions \textbf{(A3)} with $\gamma =1$ and  \textbf{(A4)} with $\kappa = 2$ are satisfied.  
%
%\begin{equation}\label{target_function_l1_loss}
 %  \text{med}(Y_0|X_0=x) = h^*(x), ~ for ~ x\in \R^d,  
%\end{equation}
%
%where med stands for the median. 
%
So, we assume that $\mathcal{X} = [0, 1]^d$ and consider the estimation of $h_{0}^{*} \coloneqq h^{*} \ind_{[0,1]^d}$.
The function $h^*$ can be written as $h^*=g_1\circ g_0$ with $g_0(x_1, \dots, x_d) = (m_1(x_1), \dots, m_d(x_d))^T$ and $g_1(x_1, \dots, x_d) = c_0 + \sum_{i=1}^d x_i$, where $m_i(x_i) = \big(c_i + \pi_i e^{\lambda(x_i - z_i)^2}\big) x_i$ for  $i=1, \dots, d$.
Let $\beta >0$. 
One can find $A>0$ such that $m_i|_{[0, 1]}\in \mathcal{C}^{\beta}([0, 1], A)$ for $i=1, \dots, d$. 
Also, $g_1 \in \mathcal{C}^{\delta}\big([-A, A]^d, (A+1)d + |c_0|\big)$ for any $\delta \ge 1$. 
So, we have $h^*|_{[0, 1]^d} : [0, 1]^d \rightarrow [-Ad - |c_0|, Ad + |c_0|]$ and
\begin{equation*}
 h^*|_{[0, 1]^d}\in \mathcal{G}\big(1, (d, d, 1), (1, d), (\beta, (\beta\lor1)d), (A+1)d + |c_0|\big).   
\end{equation*}
Therefore, $\phi _n = n^{-\frac{2\beta}{2\beta + 1}}$.
\end{example}
 
\subsection{Binary classification with the logistic Loss}
Consider a stationary ergodic strong mixing process ${Z _t = (X_t, Y_t), t\in \Z}$ which take values in $\mx \times \{-1, 1\} \subset \R ^d \times \R$ satisfying for all $x \in \mx$,
\begin{equation}\label{def_binary_model}
    Y_t | X_t= x \sim 2 \mathcal{B}(\eta(x)) - 1, \text{ with }\eta(x) = P(Y_t = 1|X_t = x), 
\end{equation}
where $\mathcal{B}(\eta(x))$ represents the Bernoulli distribution with parameter $\eta(x)$. We deal with on a margin-based logistic loss function given by, $\ell(y, u) = \phi(yu)$ for all $(y, u) \in \{-1, 1\} \times \R$ and with $\phi(v) = \log \big( 1 +  e ^{-v} \big)$ for all $v \in \R$. For any estimator $h : \R ^d \to \R$, its risk with respect to the loss function $\phi(\cdot)$ is defined by,
\[ R(h) = \E[\phi(Y_0 h(X_0))].   \]
In this setting, we assume that $\mx \subset \R ^d ~(d\in \N)$ is a compact set, and we focus on the deep neural network class $\mathcal{H}_{\sigma}(L_n, N_n, B_n, F_n)$ with $F_n = F > \max(\mk, 1)$. Thus, the assumptions \textbf{(A0)}-\textbf{(A3)} are satisfied.
 The target function, with respect to the $\phi$-loss, as given (see  also Lemma C.2 in \cite{zhang2024classification}), is the function $h ^* : \mx \to [-\infty, \infty]$, defined by:
\begin{equation}\label{def_target_logistic_loss_v1}
h ^*(x) =  \left\{
\begin{array}{ll}
      \infty &  \text{ if } ~ \eta(x) = 1 , \\
\log \dfrac{\eta(x)}{1 - \eta(x)} &   ~ \text{ if } ~ \eta(x) \in (0, 1), \\
-\infty &  \text{ if } ~ \eta(x) = 0. 
\end{array}
\right.      
\end{equation} 
One can see from \cite{alquier2024minimax} that, \textbf{(A4)} holds with $\kappa = 2$. 
For $\mx = [0, 1] ^d$, if $h ^{*} \in \mathcal{G}(q, \bold{d}, \bold{t}, \boldsymbol{\beta}, \mathcal{K})$, then the upper bound in (\ref{excess_risk_bound_corol01_v1}) matches, up to logarithmic factor, the lower bound establishes in Theorem $5.2$ by \cite{alquier2024minimax}. 
Hence, for the classification problem with the logistic loss, the SPDNN predictor is optimal (up to a logarithm factor) in the minimax sense, on the class of composition structured H\"older functions.

\begin{example}[Binary autoregressive model]
We consider a binary autoregressive process with exogenous covariates $( Y_t , \mx_t)_{t \in \Z}$ with values in $\{-1 , 1 \} \times \R^{d_x}$ ($d_x \in \N$) satisfying:
\begin{equation}\label{binary_INGARCH_model}
Y_t| \mathcal{F} _{t-1} \sim 2 \mathcal{B}(\eta _t)  \text{ with } 2\eta _t - 1 = \E[Y_t | \mathcal{F} _{t-1}] = f(Y _{t-1}, Y _{t-2}, \cdots, Y _{t - p}) + g(\mx _{t-1}), 
\end{equation}
where $p \in \N$, $\mathcal{F} _{t-1} = \sigma(Y _{t-1}, \cdots , Y _{t-p}; \mx _{t-1})$ denotes the $\sigma$-field generated by the past at time $t-1$, $f, g$ are two measurable functions such that, for all $(y_1, \cdots, y_p) \in \{-1, 1 \}^p$ and $ x \in \R ^{d_x}$, $f(y_1, \cdots, y_p) + g(x) \in [-1, 1]$. 
Here, $\eta _t = P(Y _t = 1|\mathcal{F} _{t-1})$ where $\mathcal{B}(\eta _t)$ is the Bernoulli distribution with parameter $\eta_t$ and $(\mx_t)_{t \in \Z}$ is an exogenous covariate which takes values in $\R^{d_x}$ and assumed to be stationary and ergodic. 
Under some Lipschitz-type conditions on $f$, it holds from Theorem 3.3 in \cite{aknouche2021count} that, there exists a solution $(Y _t)_{t \in \Z}$ of the model (\ref{binary_INGARCH_model}) which is strongly mixing with geometrically convergence rate.
So, for the deep learning problem from $( Y_t , X_t)_{t \in \Z}$  with $X_ t = (Y _{t-1}, \cdots, Y _{t-d}) $ and with the logistic loss function, the second item of Corollary \ref{corol1} is applied with $\kappa=2$. 
Also, by encoding $(Y_t)_{t \in \Z}$ into 0, 1 valued process, one can set $\mx=[0,1]^d$ and the second item of Corollary \ref{corol01_v1} can be applied with $\kappa=2$. That is, the SPDNN predictor studied above  is optimal (up to a logarithmic factor) in the minimax sense on the class $\mathcal{G}(q, \bold{d}, \bold{t}, \boldsymbol{\beta}, \mathcal{K})$ for the binary classification of strongly mixing processes.     
\end{example}

\section{Proofs of the main results}\label{prove}
%
%Let us Proof some useful Lemma in the sequel
The following lemma will be useful in the sequel.
\begin{lem}\label{Lemma_bernstein_inequa}
Assume that $\{ U_i \}_{i \ge 1}$ is an exponentially strongly mixing sequence drawn from $\mathcal{Z}$, and let $\mathcal{G}$ be a set of functions on $\mathcal{Z}$ such that, for some $C_1, C_2 \ge 0$, $|g(U_1) - \E[g(U_1)]| \leq C_1$ $a.s.$ and $\E[g(U_1)] ^2 \leq C_2 \E[g(U_1)]$ for each $g \in \mathcal{G}$. Then for every $\varepsilon > 0$, and $0 < u \leq 1$, there holds
\begin{equation}
P\Big\{ \sup_{g \in \mathcal{G}} \dfrac{\E[g(U_1)] - (1/n) \sum_{i = 1} ^n g(U_i)}{\sqrt{ E[g(U_1)] + \varepsilon }} \ge 4u \sqrt{\varepsilon} \Big\} \lesssim \mathcal{N}(\mathcal{G}, u\varepsilon) \cdot \exp \Bigg( -\dfrac{C_{3} u^2 \varepsilon \big(n/(\log n )^2 \big) }{3 C_2 +  C_1 } \Bigg),
\end{equation}
for some constant $C_3 > 0$.
\end{lem}

\medskip

\medskip

\noindent
\textbf{Proof of Lemma \ref{Lemma_bernstein_inequa} }
Set $m :=  \mathcal{N}(\mathcal{G}, u\varepsilon)$ the $ u\varepsilon$- covering number of $\mathcal{G}$, then there exists $m$ balls $\{B_j \}_{j=1}^m$ covering $\mathcal{G}$, for which
$B_j = \{g \in \mathcal{G}: \|g_j - g\|_{\infty} \leq u\varepsilon\}$. Then for each $g \in B_j$, since $\E[g(U_1)] \geq 0$, we have
\begin{align}\label{proof_of_lemma_v1}
\nonumber \dfrac{| (1/n) \sum_{i = 1} ^n \E[g(U_i)] - (1/n) \sum_{i = 1} ^n \E[g_j(U_i)] |}{\sqrt{ \E[g(U_1)] + \varepsilon }} & \leq \dfrac{\E|g(U_1) - g_j(U_1) |}{\sqrt{ \E[g(U_1)] + \varepsilon }}   \leq \dfrac{\| g- g_j \|_{\infty}}{\sqrt{\E[g(U_1)] + \varepsilon }} \\
& \leq \dfrac{u\varepsilon}{\sqrt{\E[g(U_1)] + \varepsilon }} \leq u\sqrt{\varepsilon}.
\end{align}
Therefore, for any $j = 1, \cdots, m$,
\begin{align}\label{proof_of_lemma_v2}
\nonumber & P\Bigg\{ \sup_{g \in B_j} \dfrac{\E[g(U_1)] - (1/n) \sum_{i = 1} ^n g(U_i)}{\sqrt{\E[g(U_1)] + \varepsilon }} \ge 4u \sqrt{\varepsilon} \Bigg\} \\
\nonumber & = P\Bigg\{ \sup_{g \in B_j} \Bigg[\dfrac{\E[g(U_1)] - \E[g_j(U_1)]}{\sqrt{ E[g(U_1)] + \varepsilon }} +  \dfrac{\E[g_j(U_1)] - (1/n) \sum_{i = 1} ^n g_j(U_i)}{\sqrt{ E[g(U_1)] + \varepsilon }} \\
\nonumber &  + \dfrac{(1/n) \sum_{i = 1} ^n g_j(U_i) - (1/n) \sum_{i = 1} ^n g(U_i)}{\sqrt{ E[g(U_1)] + \varepsilon }} \Bigg] \ge 4u \sqrt{\varepsilon} \Bigg\}  \\
\nonumber & \leq  P\Bigg\{\dfrac{\E[g_j(U_1)] - (1/n) \sum_{i = 1} ^n g_j(U_i)}{\sqrt{ \E[g_j(U_1)] + \varepsilon }} \ge u \sqrt{\varepsilon} \Bigg\}  \\
\nonumber & \leq  P\Bigg\{\E[g_j(U_1)] - (1/n) \sum_{i = 1} ^n g_j(U_i) \ge u\sqrt{ \varepsilon \big( \E[g_j(U_1)]/\varepsilon + 1 \big) } \Bigg\} \\
\nonumber & \leq  P\Bigg\{\dfrac{\E[g_j(U_1)] - (1/n) \sum_{i = 1} ^n g_j(U_i)}{\sqrt{ \E[g_j(U_1)] + \varepsilon }} \ge u \sqrt{\varepsilon} \Bigg\}  \\
 & \leq  P\Bigg\{\dfrac{1}{n} \sum_{i = 1} ^n \Big( \E[g_j(U_1)] - g_j(U_i) \Big) \ge u\sqrt{ \varepsilon \big( \E[g_j(U_1)]/\varepsilon + 1 \big)} \Bigg\}.
\end{align}
To bound the right-hand side of (\ref{proof_of_lemma_v2}), we use the exponential inequality derived in Subsection 3.3.3 of \cite{hang2016learning}, which is based on the result from \cite{merlevede2009bernstein}.
For notational simplicity, we set $w = \E[g_j(U_1)]$ in the sequel. Thus, from \cite{hang2016learning} and \cite{merlevede2009bernstein}, there exist a constant $C_3>0$ independent of $n$, and $n_0 \in \N$ such that for any $n \ge n_0$, we have

\begin{align}\label{proof_of_lemma_v4}
 \nonumber P\Bigg\{\dfrac{1}{n} \sum_{i = 1} ^n \Big( w - g_j(U_i) \Big) \ge u\sqrt{ \varepsilon \big( w/\varepsilon + 1 \big)} \Bigg\} & \lesssim \exp \Bigg( -\dfrac{ C_{3} u^2 \varepsilon \big( w/\varepsilon + 1 \big) \big(n/(\log n) ^2 \big)}{3 C_2 w + u\sqrt{ \varepsilon \big( w/\varepsilon + 1 \big)} C_1 } \Bigg)  \\
\nonumber & \lesssim  \exp \Bigg( -\dfrac{C_{3} u^2 \varepsilon^2 \big( w/\varepsilon + 1 \big) \big(n/(\log n) ^2 \big) }{3 C_2 \varepsilon (w/\varepsilon + 1) + \varepsilon \big( w/\varepsilon + 1 \big) C_1 } \Bigg) \\
& \lesssim    \exp \Bigg( -\dfrac{ C_{3} u^2 \varepsilon \big(n/ (\log n) ^2 \big) }{3 C_2 +  C_1 } \Bigg),
\end{align}
Thus, we have
\begin{align}\label{proof_of_lemma_v5}
\nonumber P\Bigg\{ \sup_{g \in \mathcal{G}} \dfrac{\E[g(U_1)] - (1/n) \sum_{i = 1} ^n g(U_i)}{\sqrt{ E[g(U_1)] + \varepsilon }} \ge 4u \sqrt{\varepsilon} \Bigg\} & \leq P\Bigg\{\bigcup_{j = 1}^m \sup_{g_j \in B_j} \dfrac{\E[g(U_1)] - (1/n) \sum_{i = 1} ^n g(U_i)}{\sqrt{ E[g(U_1)] + \varepsilon }} \ge 4u \sqrt{\varepsilon} \Bigg\} \\
\nonumber & \leq \sum_{j = 1}^m P\Bigg\{ \sup_{g_j \in B_j} \dfrac{\E[g(U_1)] - (1/n) \sum_{i = 1} ^n g(U_i)}{\sqrt{ E[g(U_1)] + \varepsilon }} \ge 4u \sqrt{\varepsilon} \Bigg\}  \\
\nonumber & \leq  m \cdot P\Bigg\{ \sup_{g_j \in B_j} \dfrac{\E[g(U_1)] - (1/n) \sum_{i = 1} ^n g(U_i)}{\sqrt{ E[g(U_1)] + \varepsilon }} \ge 4u \sqrt{\varepsilon} \Bigg\}   \\
\nonumber &   \lesssim \mathcal{N}(\mathcal{G}, u\varepsilon) \cdot  \exp \Bigg( -\dfrac{C_{3} u^2 \varepsilon \big(n/(\log n )^2 \big) }{3 C_2 +  C_1 } \Bigg).
\end{align}

\subsection{Proof of Theorem \ref{thm1}}
Let $D_n = \{(X_i, Y_i)_{i=1,\cdots,n}\}$ generated from a stationary and ergodic process $\{Z_t = (X_t, Y_t), t\in \Z\}$ which takes values in $\mathcal{Z} = \mathcal{X} \times \mathcal{Y}$.
Recall that $h^{*}$ is defined in (\ref{best_pred_F}), and $\widehat{h}_n$ given in (\ref{sparse_DNNs_Estimators}) is obtained from the empirical risk minimization algorithm depending on the training sample $D_n$ and its expected excess risk is defined by:
\begin{equation}\label{proof_excess_risk_v1}
 \E[R(\widehat{h}_n) - R(h^{*}) ]  \coloneqq \E[B_{1, n}] +  \E[B_{2, n}],
 \end{equation}
where,
\[
\begin{array}{llll}
  B_{1, n}   & =   [R(\widehat{h}_n) - R(h^{*})]  - 2 [ \widehat{R}_n (\widehat{h}_n)  - \widehat{R}_n (h^{*}) ] - 2 J_n(\widehat{h}_n); 
  \\
 B_{2, n}   &  =   2[\widehat{R}_n (\widehat{h}_n) - \widehat{R}_n (h^{*})] + 2 J_n(\widehat{h}_n).
\end{array}
\]
In the sequel, we will derive a bound of $\E[B_{1, n}]$ and $\E[B_{2, n}]$.

\medskip

\noindent
\textbf{Step 1: Bounding the first term in the right-hand side of (\ref{proof_excess_risk_v1})}.

\medskip

\noindent
Let $\rho > 1/ n^{(\alpha)}$ and $L_n, N_n, B_n, F_n = F > 0$, fulfill the conditions in Theorem \ref{thm1}. We set in the following 
\begin{equation}\label{proof_penalty_neural_net_archi}
    \mathcal{H}_{\sigma, n} \coloneqq \mathcal{H}_{\sigma}(L_n, N_n, B_n, F). 
\end{equation}
Let us define 
\begin{equation*}
 \mathcal{H}_{n, j, \rho} \coloneqq  \{ h \in \mathcal{H}_{\sigma, n}: 2^{j - 1} \ind_ {\{j \ne 0 \} } \rho  \leq J_n(h) \leq  2^{j} \rho \}.   
\end{equation*}
 Let
\[\Delta(h) (Z_0) \coloneqq \ell (h(X_0), Y_0) - \ell (h^{*}(X_0), Y_0), \text{ with } Z_0  \coloneqq (X_0, Y_0).     
\]
Thus, for $\rho > 1/ n^{(\alpha)}$ we have
\begin{align}\label{proof_bound_B1n_v1}
\nonumber & P(B_{1, n} > \rho)  = P\Big( \E [\ell ( \widehat{h}_n (X_0), Y_0)] - \E[\ell(h^{*} (X_0), Y_0)] - \dfrac{2}{n} \sum_{i= 1}^n [\ell (\widehat{h}_n (X_i), Y_i) -  \ell((h^{*} (X_i), Y_i) ] - 2 J_n(\widehat{h}_n) >  \rho \Big)
\\
\nonumber & \leq P\Big( \exists h \in \mathcal{H}_{\sigma, n}: \E [\ell( h (X_0), Y_0)] - \E[ \ell (h^{*} (X_0), Y_0)] - \dfrac{1}{n} \sum_{i= 1}^n [\ell ( h (X_i), Y_i) -  \ell (h^{*} (X_i), Y_i)] > \dfrac{1}{2} \Big(\E [\ell( h(X_0), Y_0)] 
\\
\nonumber & \hspace{11.4cm} - \E[\ell (h^{*} (X_0), Y_0)]   + 2 J_n(h) + \rho \Big) \Big)  
\\
\nonumber &  \leq P \Big( \exists h \in \mathcal{H}_{\sigma, n}: \dfrac{1}{n} \sum_{i= 1}^{n} \Big[\E[\Delta (h) (Z_i)] -\Delta (h) (Z_i) \Big] > \dfrac{1}{2}\Big(\rho + 2 J_n(h) + \E[\Delta (h) (Z_0)] \Big)\Big) \\
\nonumber & \leq P\Bigg(   \sup_{ h \in \mathcal{H}_{\sigma, n}} \dfrac{ \dfrac{1}{n} \sum_{i= 1}^{n} \Big[\E[\Delta (h) (Z_i)] -\Delta (h) (Z_i) \Big]  }{ \rho + 2 J_n(h) + \E[\Delta (h) (Z_0)]   }  > \dfrac{1}{2} \Bigg) \\
\nonumber & \leq \sum_{j=1}^ {\infty}P\Bigg( \sup_{ h \in \mathcal{H}_{n, j, \rho}} \dfrac{ \dfrac{1}{n} \sum_{i= 1}^{n} \Big[\E[\Delta (h) (Z_i)] -\Delta (h) (Z_i) \Big]  }{ \rho + 2 J_n(h) + \E[\Delta (h) (Z_0)]   }  > \dfrac{1}{2}  \Bigg)  \\
& \leq \sum_{j=1}^ {\infty}P\Bigg( \sup_{ h \in \mathcal{H}_{n, j, \rho}} \dfrac{ \dfrac{1}{n} \sum_{i= 1}^{n} \Big[\E[\Delta (h) (Z_i)] -\Delta (h) (Z_i) \Big]  }{ 2^ j \rho + \E[\Delta (h) (Z_0)]   }  > \dfrac{1}{2}  \Bigg).
\end{align}
Indeed,
\[\text{for } \rho  >0,  j \geq 0 \text{ and } h \in \mathcal{H}_{n, j, \rho},  ~ 2 J_n(h) \geq 2^j \rho.  
\]
Since $\E[\Delta (h) (Z_0)] = \E[\ell (h(X_0), Y_0)] - \E[\ell (h^{*}(X_0), Y_0)] \ge 0$, according to (\ref{proof_bound_B1n_v1}), we get,
\begin{align}\label{proof_bound_B1n_v2}
\nonumber P(B_{1, n} > \rho) &\leq \sum_{j=1}^ {\infty}P\Bigg( \sup_{ h \in \mathcal{H}_{n, j, \rho}} \dfrac{ \dfrac{1}{n} \sum_{i= 1}^{n} \Big[\E[\Delta (h) (Z_i)] -\Delta (h) (Z_i) \Big]  }{ \sqrt{ 2^ j \rho + \E[\Delta (h) (Z_0)]} \sqrt{2^ j \rho} }  > \dfrac{1}{2}  \Bigg)  \\
&\leq \sum_{j=1}^ {\infty}P\Bigg( \sup_{ h \in \mathcal{H}_{n, j, \rho}} \dfrac{ \dfrac{1}{n} \sum_{i= 1}^{n} \Big[\E[\Delta (h) (Z_i)] -\Delta (h) (Z_i) \Big]  }{ \sqrt{ 2^ j \rho + \E[\Delta (h) (Z_0)]} }  > \dfrac{1}{2}\sqrt{2^ j \rho}  \Bigg). 
\end{align}
Set
\begin{equation*}
\mathcal{G}_{n, j, \rho} \coloneqq \{ \Delta (h): \R^d \times \mathcal{Y} \rightarrow  \R: h \in \mathcal{H}_{n, j, \rho} \}.
\end{equation*}
Let $g(x, y) = \Delta(h)(x, y)$ for all $h \in \mathcal{H}_{n, j, \rho}$. 
Recall the conditions $\|h\|_{\infty} \leq F$, $\|h^{*}\|_{\infty} \leq \mathcal{K}$, and the assumption that the loss function $\ell$ is $\mathcal{K}_{\ell}$-Lipschitzian. Thus, we have, a.s.
\begin{equation} \label{proof_bound_of_g}
|g(Z_0)| = |\Delta(h)(Z_0)| \leq \mathcal{K}_{\ell}(F + \mathcal{K}), ~ \text{with} ~ Z_0 = (X_0, Y_0).
\end{equation}
Hence,  
\begin{equation}\label{proof_bound_of_g_carre}
 \var (g(Z_0)) \leq \E[g(Z_0)^ 2] \leq \mathcal{K}_{\ell}(F + \mathcal{K})\E[g(Z_0)]. 
\end{equation}
According to (\ref{proof_bound_of_g}), we get,
\begin{equation}\label{proof_bound_of_the_process}
|\E [g(Z_0)] - g(Z_0)| \leq 2\mathcal{K}_{\ell}(F + \mathcal{K}).
\end{equation}
Let $j\in \N$. From (\ref{proof_bound_of_g_carre}), (\ref{proof_bound_of_the_process}) and the Bernstein inequality for strong mixing processes, see Lemma $3$ in \cite{xu2008learning}, we have, 
\begin{align}\label{equ_b}
&\nonumber P\Bigg( \sup_{ h \in \mathcal{H}_{n, j, \rho}} \dfrac{ \dfrac{1}{n} \sum_{i= 1}^{n} \Big[\E[\Delta (h) (Z_i)] -\Delta (h) (Z_i) \Big]  }{ \sqrt{ 2^ j \rho + \E[\Delta (h) (Z_0)]} }  > \dfrac{1}{2}\sqrt{2^ j \rho}  \Bigg)  = P\Bigg( \sup_{ h \in \mathcal{G}_{n, j, \rho}} \dfrac{ \dfrac{1}{n} \sum_{i= 1}^{n} \Big[\E[g (Z_i)] - g (Z_i) \Big] }{ \sqrt{ 2^ j \rho + \E[g (Z_0)]} }  > \dfrac{1}{2}\sqrt{2^ j \rho}  \Bigg) \\ 
\nonumber & \leq \mathcal{N} \Big(\mathcal{G}_ {n, j, \rho}, \dfrac{2^ j \rho}{8}, \|\cdot\|_{\infty} \Big) (1 + 4e^ {-2} \overline{\alpha}) \exp \Big[- \dfrac{\dfrac{2^ j \rho}{64} n^ {(\alpha)}}{2\mathcal{K}_{\ell}(F + \mathcal{K}) + 4\mathcal{K}_{\ell}(F + \mathcal{K})/3}\Big]   \\
&\leq \mathcal{N} \Big(\mathcal{G}_ {n, j, \rho}, \dfrac{2^ j \rho}{8}, \|\cdot\|_{\infty} \Big) (1 + 4e^ {-2} \overline{\alpha}) \exp \Big[- \dfrac{3(2^ j \rho)n^ {(\alpha)}}{ 640\mathcal{K}_{\ell}(F + \mathcal{K}) }\Big],       
\end{align}
%
%
%where $\mathcal{N} \Big(\mathcal{G}_ {n, j, \rho}, \varepsilon \Big)$ denotes the $\varepsilon$-covering number of $ \mathcal{G}_ {n, j, \rho}$ see, (\ref{epsi_covering_number}) in Section \ref{asump}.
%
%The condition imposed on the function $\pi_{\lambda_n, \tau_n}$, implies $\pi_{\lambda_n, \tau_n}(x) \leq \lambda_n \ind_{x \neq 0}$ for all $x \ge 0$ and thus $ J_n(h) \leq \lambda_n \| \theta(h)\|_0$ for any neural network $h$. Thus, we get, 
%
%One can easily see that,
%
%\begin{equation}\label{inclusion}
 %\mathcal{H}_{n, j, \rho}  \subset \Big\{  h \in \mathcal{H}_{\sigma}(L_{n}, N_{n}, B_{n}, F, \dfrac{2^j \rho}{\lambda_n}): \| \theta(h) \|_0 \leq \frac{2^j \rho}{\lambda_n} \Big\}. 
 %\end{equation}
%
%
Let $ \Delta(h_1), ~  \Delta(h_2) \in \mathcal{G}_{n, j, \rho}, ~ \text{and} ~ (x, y) \in \R^d \times \R$, we have
\begin{align*}
\nonumber \| \Delta(h_1) (x, y) -  \Delta(h_2) (x, y) \|_{\infty} & = | \ell (h_1(x), y) - \ell (h_2(x), y)| 
\\
& \leq  \mathcal{K}_{\ell}|h_1 (x) - h_2 (x)|.
\end{align*}
Set $\varepsilon = \dfrac{2^ j \rho}{8}$. 
According to the condition on $\tau_n$ (in the statement of the theorem), we have $\varepsilon/\mathcal{K}_{\ell} > \tau_n (L_n + 1)((N_n + 1) B_n)^{L_n +1}$. From \cite{ohn2022nonconvex}, see also Theorem $4.1$ in \cite{kengne2024sparse}, the following inequality holds:
\begin{align}\label{equa_covering_number_Gn}
\mathcal{N} \Big(\mathcal{G}_{n, j, \rho}, \varepsilon, \| \cdot\|_{\infty} \Big)  & \leq \mathcal{N} \Big(\mathcal{H}_{n, j, \rho}, \dfrac{\varepsilon}{\mathcal{K}_{\ell}}, \| \cdot\|_{\infty} \Big).
\end{align} 
For all $L, N, B, F, S \ge 0$, set 
\[\widetilde{\mathcal{H}}_{\sigma} (L, N, B, F, S) \coloneqq \{h \in \mathcal{H}_{\sigma} (L, N, B, F): J_n(h) \leq \lambda _n S \}.
\]
We have for all $j \in \N$ and $\rho > 0$
\[
\mathcal{H}_{n, j, \rho} \subset \widetilde{\mathcal{H}}_{\sigma} (L, N, B, F, \dfrac{2^j \rho}{\lambda _n}). 
\]
Hence, from Lemma E.6 in \cite{kurisu2025adaptive} and according to (\ref{equa_covering_number_Gn}), it holds that, for all $\varepsilon \in \big( \mathcal{K} _{\ell} \tau _n (L _n + 1)((N _n + 1)B _n)^{L _n + 1}, 1\big)$, 
\begin{align}\label{proof_cover_number_v1}
\nonumber \mathcal{N} \Big(\mathcal{G}_{n, j, \rho}, \varepsilon, \|\cdot\|_{\infty} \Big)  & \leq \mathcal{N} \Big(\mathcal{H}_{n, j, \rho}, \dfrac{\varepsilon}{\mathcal{K}_{\ell}}, \| \cdot\|_{\infty} \leq \mathcal{N}\Big( \widetilde{\mathcal{H}}_{\sigma}(L_{n},N_{n}, B_{n}, F, \frac{2^j \rho}{\lambda_n}),  \frac{\varepsilon}{\mathcal{K}_{\ell}}, \|\cdot\|_{\infty} \Big)\\
& \leq \exp\Bigg[ 2 \frac{2^j \rho}{\lambda_n}(L_n + 1) \log \Bigg(\frac{(L_n + 1)(N_n + 1)B_n}{\dfrac{\varepsilon}{\mk _{\ell}} - \tau_n (L_n + 1)((N_n + 1) B_n)^{L_n +1}} \Bigg) \Bigg].
\end{align} 
According to (\ref{proof_bound_B1n_v2}), (\ref{equ_b}) and (\ref{proof_cover_number_v1}), it holds that,
\begin{align}\label{proof_proba_bound_B1n_v3}
 \nonumber & P\big( B_{1, n} > \rho \big) \\
 &\leq (1 + 4e^{-2}\overline{\alpha})\sum _{j = 1}^{\infty} \exp\Bigg[ 2 \frac{2^j \rho}{\lambda_n}(L_n + 1) \log \Bigg(\frac{(L_n + 1)(N_n + 1)B_n}{(2^j\rho)/(8\mathcal{K}_{\ell}) - \tau_n (L_n + 1)((N_n + 1) B_n)^{L_n +1}} \Bigg) - \dfrac{3(2^ j \rho)n^ {(\alpha)}}{ 640\mathcal{K}_{\ell}(F + \mathcal{K}) } \Bigg].
\end{align}

\medskip

\noindent
\textbf{Case 1}: $\rho > 1$. Recall the condition
\[ \tau_ n \leq \dfrac{1}{16\mathcal{K}_ {\ell} (L_ n + 1)((N_ n + 1) B_ n)^{L_ n +1}}.     
\]
Since $2^j\rho > 1$, we have,
\begin{align*} %\label{equa_taun_v1}
\nonumber \dfrac{2^j \rho}{8 \mathcal{K}_{\ell}} > \dfrac{1}{8 \mathcal{K}_{\ell}} & \Longrightarrow  \dfrac{2^j \rho}{8 \mathcal{K}_{\ell}} - \dfrac{1}{16 \mathcal{K}_{\ell}} > \dfrac{2^j \rho}{8 \mathcal{K}_{\ell}} - \dfrac{2^j \rho}{16 \mathcal{K}_{\ell}} = \dfrac{2^j \rho}{16 \mathcal{K}_{\ell}}             \\
& \Longrightarrow \dfrac{(L_n + 1)(N_n + 1)B_n}{\dfrac{2^j \rho}{8 \mathcal{K}_{\ell}} - \dfrac{1}{16 \mathcal{K}_{\ell}}} < \dfrac{16 \mathcal{K}_{\ell} (L_n + 1)(N_n + 1)B_n}{ 2^j \rho}.
\end{align*}
Therefore, in addition to (\ref{proof_proba_bound_B1n_v3}), gives,
\begin{align}\label{proof_proba_bound_B1n_v4}
P\big( B_{1, n} > \rho \big) \leq (1 + 4e^{-2}\overline{\alpha})\sum _{j = 1}^{\infty} \exp\Bigg[ 2 \frac{2^j \rho}{\lambda_n}(L_n + 1) \log \big(16\mathcal{K}_ {\ell} (L_n + 1)(N_n + 1) B_n \big) - \dfrac{3(2^ j \rho)n^{(\alpha)}}{ 640\mathcal{K}_{\ell}(F + \mathcal{K}) } \Bigg].
\end{align}
Recall the following  assumptions: 
\begin{equation}\label{equa_network_archi}
 L_n \asymp \log\big(n^{(\alpha)}\big), N_n \lesssim \big(n^{(\alpha)}\big)^{\nu_1}, B_n \lesssim \big(n^{(\alpha)} \big)^{\nu_2}, \lambda_n \asymp \frac{\big( \log\big(n^{(\alpha)}\big) \big)^{\nu_3}}{n^{(\alpha)}}, ~ \text{for some} ~ ,\nu_1, \nu_2 >0, \nu_3 > 2.   
\end{equation}
According to (\ref{equa_network_archi}), we get,
\begin{align}\label{proof_proba_bound_B1n_v5}
\nonumber & P\big( B_{1, n} > \rho \big)  \\
\nonumber &\lesssim \sum _{j = 1}^{\infty} \exp\Bigg( 2 \frac{2^j \rho n^{(\alpha)}}{ \big( \log \big( n^ {(\alpha)}\big) \big)^ {\nu_3} } \big(\log \big( n^ {(\alpha)}\big) + 1 \big) \log \Big( 16\mathcal{K}_ {\ell} \big( \log \big( n^ {(\alpha)} \big) + 1 \big) \big( \big( n^{(\alpha)} \big)^ {\nu_1} + 1 \big) \big( n^ {(\alpha)} \big)^ {\nu_2} \Big) - \dfrac{3(2^ j \rho)n^ {(\alpha)}}{ 640\mathcal{K}_{\ell}(F + \mathcal{K}) } \Bigg)  \\
\nonumber &\lesssim \sum_ {j = 1}^ {\infty} \exp \Bigg[ \dfrac{3(2^ j \rho)n^ {(\alpha)}}{ 640\mathcal{K}_{\ell}(F + \mathcal{K}) }  \\
 & \hspace{0.2cm} \times \Bigg(\frac{1280 \mathcal{K}_{\ell}(F + \mathcal{K})}{ 3\big( \log \big( n^ {(\alpha)}\big) \big)^ {\nu_3} } \big(\log \big( n^ {(\alpha)}\big) + 1 \big) \log \Big( 16\mathcal{K}_ {\ell} \big( \log \big( n^ {(\alpha)} \big) + 1 \big) \big( \big( n^{(\alpha)} \big)^ {\nu_1} + 1 \big) \big( n^ {(\alpha)} \big)^ {\nu_2} \Big) -  1 \Bigg) \Bigg]. 
\end{align}
Since $\nu_3 > 2$, we have,
\begin{align*}
\frac{1280 \mathcal{K}_{\ell}(F + \mathcal{K})}{ 3\big( \log \big( n^ {(\alpha)}\big) \big)^ {\nu_3} } \big(\log \big( n^ {(\alpha)}\big) + 1 \big) \log \Big( 16\mathcal{K}_ {\ell} \big( \log \big( n^ {(\alpha)} \big) + 1 \big) \big( \big( n^{(\alpha)} \big)^ {\nu_1} + 1 \big) \big( n^ {(\alpha)} \big)^ {\nu_2} \Big) \limiten 0.    
\end{align*}

\medskip

\noindent
Then, there exists $ n_1 \coloneqq n_1(\mathcal{K}_{\ell}, F, \mathcal{K}, \nu_1, \nu_2, \nu_3) \in \N$ such that, for any $ n > n_1$, we have,
\begin{align*}
\frac{1280 \mathcal{K}_{\ell}(F + \mathcal{K})}{ 3\big( \log \big( n^ {(\alpha)}\big) \big)^ {\nu_3} } \big(\log \big( n^ {(\alpha)}\big) + 1 \big) \log \Big( 16\mathcal{K}_ {\ell} \big( \log \big( n^ {(\alpha)} \big) + 1 \big) \big( \big( n^{(\alpha)} \big)^ {\nu_1} + 1 \big) \big( n^ {(\alpha)} \big)^ {\nu_2} \Big) < \dfrac{1}{2}.    
\end{align*}
Hence, (\ref{proof_proba_bound_B1n_v5}) becomes,
\begin{align}\label{proof_proba_bound_B1n_v6}
P\big( B_{1, n} > \rho \big) 
&\lesssim \sum_ {j = 1}^ {\infty} \exp \Bigg[- \dfrac{3(2^ j \rho)n^ {(\alpha)}}{ 1280\mathcal{K}_{\ell}(F + \mathcal{K}) } \Bigg]. 
\end{align}
In the following, we set:
\[\beta_n \coloneqq \rho \dfrac{ 3n^ {(\alpha)} }{ 1280\mathcal{K}_{\ell}(F + \mathcal{K}) } \geq \dfrac{ 3n^ {(\alpha)} }{ 1280\mathcal{K}_{\ell}(F + \mathcal{K}) } \limiten \infty.   
\]
Thus, for sufficiently large $n$, $\exp(-\beta_n) < \frac{1}{2}$.
Once can easily get,  
\begin{equation} \label{boundbetan}
\sum_{j =0}^{\infty} \exp(-\beta_n 2^j) \leq e^{-\beta_n} \sum_{j = 0}^{\infty} \big(e^{-\beta_n} \big)^j \leq e^{-\beta_n} \Big( \dfrac{1}{1 - e^{-\beta_n}} \Big) \leq 2e^{-\beta_n}.
\end{equation}
So, from (\ref{proof_proba_bound_B1n_v6}), we have
\begin{equation}\label{proof_proba_bound_B1n_v7}
P(B_{1, n} > \rho) \lesssim \exp\Big(-\rho \dfrac{ 3n^ {(\alpha)} }{ 1280\mathcal{K}_{\ell}(F + \mathcal{K}) } \Big).
\end{equation}
(\ref{proof_proba_bound_B1n_v7}) fulfills for all $\alpha > 1$. Thus, we get,
\begin{align}\label{proof_proba_bound_B1n_v8}
\nonumber \displaystyle\int_{1}^{\infty} P(B_{1, n} > \rho) d\rho  &\lesssim \displaystyle\int_{1}^{\infty} \exp\Big(-\rho \dfrac{ 3n^ {(\alpha)} }{ 1280\mathcal{K}_{\ell}(F + \mathcal{K}) } \Big) d\rho   \\
&\lesssim \frac{1}{n^{(\alpha)}}\exp\Big(-\dfrac{ 3n^ {(\alpha)} }{ 1280\mathcal{K}_ {\ell}( F + \mathcal{K} ) } \Big) \lesssim \frac{1}{n^{(\alpha)}}.
\end{align}
\textbf{Case 2}: $1/n^{(\alpha)} < \rho < 1$.
Recall the condition
\[ \tau_n  \leq  \dfrac{1/n^{(\alpha)}}{16\mathcal{K}_{\ell}(L_n + 1) ((N_n + 1)B_n)^{L_n + 1}}.   
\]
Since $\rho > 1/n^{(\alpha)}$, we have also,
\begin{equation*}
\tau_n \leq \dfrac{\rho}{16\mathcal{K}_{\ell}(L_n + 1) ((N_n + 1)B_n)^{L_n + 1}}.   
\end{equation*}
In addition to (\ref{proof_proba_bound_B1n_v3}), we obtain,
\begin{align*}
P\big( B_{1, n} > \rho \big) \lesssim \sum_ {j = 1}^ {\infty} \exp\Bigg( 2 \frac{2^j \rho}{\lambda_n}(L_n + 1) \log \Big(\frac{(L_n + 1)(N_n + 1)B_n}{\big( \rho/(8\mathcal{K}_{\ell}) \big)(2^ j - 1/2)} \Big) - \dfrac{3(2^ j \rho)n^ {(\alpha)}}{ 640\mathcal{K}_{\ell}(F + \mathcal{K}) } \Bigg).
\end{align*}
Since, $\dfrac{1}{\big( \rho/(8\mathcal{K}_{\ell}) \big)(2^ j - 1/2)} \leq \dfrac{16\mathcal{K}_{\ell}}{\rho} $ and according to the condition $1/\rho < n^{(\alpha)}$, we have,
\begin{align}\label{proof_proba_bound_B1n_v9}
\nonumber P\big( B_{1, n} > \rho \big)  &\lesssim \sum_ {j = 1}^ {\infty} \exp\Bigg( 2 \frac{2^j \rho}{\lambda_n}(L_n + 1) \log \Big(\frac{16\mathcal{K}_ {\ell}(L_n + 1)(N_n + 1)B_n}{\rho} \Big) - \dfrac{3(2^ j \rho)n^ {(\alpha)}}{ 640\mathcal{K}_{\ell}(F + \mathcal{K}) } \Bigg) \\
&\lesssim \sum_ {j = 1}^ {\infty} \exp\Bigg( 2 \frac{2^j \rho}{\lambda_n}(L_n + 1) \log \Big(16\mathcal{K}_ {\ell} n^{(\alpha)}(L_n + 1)(N_n + 1)B_n  \Big) - \dfrac{3(2^ j \rho)n^ {(\alpha)}}{ 640\mathcal{K}_{\ell}(F + \mathcal{K}) } \Bigg).
\end{align}
In addition to (\ref{equa_network_archi}), we get,
\begin{align}\label{proof_proba_bound_B1n_v10}
\nonumber  P\big( B_{1, n} > \rho \big)
\nonumber & \lesssim \sum_ {j = 1}^ {\infty} \exp\Bigg[2\frac{2^j \rho n^{(\alpha)}}{\big( \log\big(n^{(\alpha)}\big) \big)^{\nu_3}}\big(\log\big(n^{(\alpha)}\big) + 1 \big)\log\Big(16n^ {(\alpha)}\mathcal{K}_ {\ell} \big(\log\big( n^ {(\alpha)} \big) + 1 \big) \big( \big( n^ {(\alpha)} \big)^ {\nu_1} + 1 \big)\big(n^ {(\alpha)}\big)^ {\nu_2} \Big)  \\
\nonumber & \hspace{12cm} - \dfrac{3(2^ j \rho)n^ {(\alpha)}}{ 640\mathcal{K}_{\ell}(F + \mathcal{K}) }  \Bigg] \\
\nonumber &  \lesssim \sum_ {j=1}^ {\infty} \exp \Bigg[ \dfrac{3(2^ j \rho)n^ {(\alpha)}}{ 640\mathcal{K}_ {\ell}(F + \mathcal{K}) }    \\
 & \hspace{0cm} \times \Bigg(\frac{1280\mathcal{K}_{\ell}(F + \mathcal{K})}{3\big( \log\big(n^{(\alpha)}\big) \big)^{\nu_3}}\big(\log\big(n^{(\alpha)}\big) + 1 \big)\log\Big(16n^ {(\alpha)} \mathcal{K}_ {\ell} \big( \log \big( n^ {(\alpha)} \big) + 1 \big) \big(\big(n^{(\alpha)}\big)^{\nu_1} + 1 \big)\big(n^{(\alpha)}\big)^{\nu_2} \Big) - 1 \Bigg)  \Bigg]. 
\end{align}
Recall that $\nu_3 > 2$ see (\ref{equa_network_archi}), thus, we have,

\medskip

\begin{align*}
\frac{1280\mathcal{K}_{\ell}(F + \mathcal{K})}{3\big( \log\big(n^{(\alpha)}\big) \big)^{\nu_3}}\big(\log\big(n^{(\alpha)}\big) + 1 \big)\log\Big(16n^ {(\alpha)} \mathcal{K}_ {\ell} \big( \log \big( n^ {(\alpha)} \big) + 1 \big) \big(\big(n^{(\alpha)}\big)^{\nu_1} + 1 \big)\big(n^{(\alpha)}\big)^{\nu_2} \Big) \limiten 0.   \end{align*}

\medskip

\noindent
Then, there exists $ n_2 \coloneqq n_2(\mathcal{K}_{\ell}, F, \mathcal{K}, \nu_1, \nu_2, \nu_3) \in \N$ such that, for any $ n > n_2$, we get,
\begin{align*}
\frac{1280\mathcal{K}_{\ell}(F + \mathcal{K})}{3\big( \log\big(n^{(\alpha)}\big) \big)^{\nu_3}}\big(\log\big(n^{(\alpha)}\big) + 1 \big)\log\Big(16n^ {(\alpha)} \mathcal{K}_ {\ell} \big( \log \big( n^ {(\alpha)} \big) + 1 \big) \big(\big(n^{(\alpha)}\big)^{\nu_1} + 1 \big)\big(n^{(\alpha)}\big)^{\nu_2} \Big) < \dfrac{1}{2}.    
\end{align*}
In addition to (\ref{proof_proba_bound_B1n_v10}), we have,
\begin{align}\label{proof_proba_bound_B1n_v11}
P\big( B_{1, n} > \rho \big) & \leq \sum _{j = 1}^{\infty} \exp \Bigg[ -\dfrac{3(2^ j \rho)n^ {(\alpha)}}{ 1280\mathcal{K}_ {\ell}(F + \mathcal{K}) } \Bigg].  
\end{align}
Let,
\[\beta_n \coloneqq \rho \dfrac{3n^ {(\alpha)}}{ 1280\mathcal{K}_ {\ell}(F + \mathcal{K}) }.     
\]
From (\ref{proof_proba_bound_B1n_v11}) and according to (\ref{boundbetan}), we have,
\begin{equation*}
P\Big(B_{1, n} > \rho \Big) \lesssim \exp\Big( -\rho \dfrac{3n^ {(\alpha)}}{ 1280\mathcal{K}_ {\ell}(F + \mathcal{K}) } \Big).
\end{equation*}
Hence, we obtain, 
\begin{align}\label{proof_proba_bound_B1n_v12}
\nonumber \displaystyle\int_{1/n^{(\alpha)}}^{1} P \Big( B_{1, n} > \rho \Big) d\rho  & \lesssim \displaystyle \int_ {1/n^{(\alpha)}}^ 1 \exp\Big( -\rho \dfrac{3n^ {(\alpha)}}{ 1280\mathcal{K}_ {\ell}(F + \mathcal{K}) } \Big) d\rho
 \\
 & \lesssim \dfrac{1}{n^{(\alpha)}} - \dfrac{1}{n^{(\alpha)}}\exp\Big( - \dfrac{3n^ {(\alpha)}}{ 1280\mathcal{K}_ {\ell}(F + \mathcal{K}) } \Big) \lesssim \dfrac{1}{n^{(\alpha)}}.
\end{align}
\textbf{Case 3}: $0 < \rho < 1/n^{(\alpha)} < 1$.
\begin{align}\label{proof_proba_bound_B1n_v13}
\displaystyle\int_{0}^{1/n^{(\alpha)}} P(B_{1, n} > \rho) d\rho & \leq \displaystyle\int_{0}^{1/n^{(\alpha)}} d\rho \leq 1/n^{(\alpha)}.
\end{align}
Set $ n_0 = \max (n_1, n_2)$. According to (\ref{proof_proba_bound_B1n_v8}), (\ref{proof_proba_bound_B1n_v12}), (\ref{proof_proba_bound_B1n_v13}), we get, for $n > n_0$
\begin{align}\label{Bound_B1n}
  \E [B_{1, n}] &\leq \int_ 0^ {\infty} P(B_ {1, n} > \rho) d\rho  \lesssim \dfrac{1}{n^ {(\alpha)}}.
\end{align}

\medskip

\noindent 
\textbf{Step 2: Bounding the second term in the right-hand side of (\ref{proof_excess_risk_v1})}.
Let us consider $h_{n}^{\diamond} \in \mathcal{H}_{\sigma, n} \coloneqq \mathcal{H}_{\sigma}(L_n, N_n, B_n, F)$ such that  
\begin{equation*}
R (h_{n}^{\diamond})  + J_n(h_{n}^{\diamond}) \leq \underset{h \in \mathcal{H}_{\sigma, n}} {\inf} \left[ R(h)  + J_n(h) \right] + \dfrac{1}{n^{(\alpha)}}.
\end{equation*}
From the definition of $\widehat{h}_n$, it holds that,
\begin{equation*}
\widehat{R}_n (\widehat{h}_n) + J_n( \widehat{h}_n) \leq  \widehat{R}_n (h) + J_n(h), ~ \text{for all} ~ h \in \mathcal{H}_{\sigma, n}.
\end{equation*}
Hence, 
\begin{align} 
\nonumber B_{2, n} & = 2[\widehat{R}_n (\widehat{h}_n) - \widehat{R}_n (h^{*}) ] + 2 J_n(\widehat{h}_n)
  = 2[\widehat{R}_n (\widehat{h}_n) + J_n(\widehat{h}_n) - \widehat{R}_n (h_{n}^{\diamond}) ] 
 + 2 [ \widehat{R}_n (h_{n}^{\diamond}) -  \widehat{R}_n (h^{*}) ]
 \\
 \nonumber & \leq 2J_n(h_{n}^{\diamond}) + 2[ \widehat{R}_n (h_{n}^{\diamond}) - \widehat{R}_n (h^{*}) ].
\end{align}
Thus,
\begin{equation}\label{B_2_n_inf}
\E[B_{2, n}] \leq 2J_n(h_n^{\diamond}) + 2\mathcal{E}_{Z_0}(h_n^{\diamond}) \leq 2 \underset{h \in \mathcal{H}_{\sigma, n}}{\inf}\Big[ \mathcal{E}_{Z_0}(h) + J_n(h) \Big] + \dfrac{1}{n^{(\alpha)}}.
\end{equation}
In the sequel, we bounded the expected excess risk.

\medskip

\noindent
\textbf{Step 3: Bounding the expected excess risk}. 
Recall that, the expected excess risk of $\widehat{h}_n$ is written as (see (\ref{proof_excess_risk_v1}))
\[ \E\big[\mathcal{E}_{Z_0}(\widehat{h}_n) \big] = \E[R(\widehat{h}_n) - R(h^{*}) ] = \E[B_{1, n}] + \E[B_{2, n}].
\]
In addition to (\ref{Bound_B1n}) and (\ref{B_2_n_inf}), we have, for all $n > n_0$,
\begin{align}\label{proof expected_excess_risk}
\E\big[\mathcal{E}_{Z_0}(\widehat{h}_n) \big] 
  \leq 2 \Big(\underset{h \in \mathcal{H}_{\sigma}(L_n, N_n, B_n, F)} {\inf} \Big[ \mathcal{E}_{Z_0} (h)  + J_n(h) \Big]  \Big)  + \dfrac{\Xi}{n^{(\alpha)}},
\end{align}
for some constant $\Xi >0$.
Which completes the proof  of the Theorem.
\qed
\subsection{Proof of Theorem \ref{oracle_inequality_expo_strong_mixing}}
As in the proof of Theorem \ref{thm1}, consider the following decomposition of the excess risk:
\begin{equation}\label{proof_excess_risk_v1_2}
 \E[R(\widehat{h}_n) - R(h^{*}) ]  \coloneqq \E[B_{1, n}] +  \E[B_{2, n}],
 \end{equation}
where,
\[
\begin{array}{llll}
  B_{1, n}   & =   [R(\widehat{h}_n) - R(h^{*})]  - 2 [ \widehat{R}_n (\widehat{h}_n)  - \widehat{R}_n (h^{*}) ] - 2 J_n(\widehat{h}_n); 
  \\
 B_{2, n}   &  =   2[\widehat{R}_n (\widehat{h}_n) - \widehat{R}_n (h^{*})] + 2 J_n(\widehat{h}_n).
\end{array}
\]
The proof follows the same steps as in the proof of Theorem \ref{thm1}.

\medskip

\noindent
\textbf{Step 1: Bounding the first term in the right-hand side of (\ref{proof_excess_risk_v1_2})}.

\medskip

\noindent
Let $\rho > 1/ n$ and $L_n, N_n, B_n, F_n = F > 0$, fulfill the conditions in Theorem \ref{oracle_inequality_expo_strong_mixing}. Set
\begin{equation*} %\label{proof_penalty_neural_net_archi}
    \mathcal{H}_{\sigma, n} \coloneqq \mathcal{H}_{\sigma}(L_n, N_n, B_n, F) ~ \text{ and } ~ 
 \mathcal{H}_{n, j, \rho} \coloneqq  \{ h \in \mathcal{H}_{\sigma, n}: 2^{j - 1} \ind_ {\{j \ne 0 \} } \rho  \leq J_n(h) \leq  2^{j} \rho \}.   
\end{equation*}
 By going along similar lines as in Step 1 in the proof of Theorem \ref{thm1} and by applying Lemma \ref{Lemma_bernstein_inequa}, we get,
\begin{align}\label{proof_proba_bound_B1n_v3}
 \nonumber & P\big( B_{1, n} > \rho \big) \\
 &\leq \sum _{j = 1}^{\infty} \exp\Bigg( 2 \frac{2^j \rho}{\lambda_n}(L_n + 1) \log \Big(\frac{(L_n + 1)(N_n + 1)B_n}{(2^j\rho)/(8\mathcal{K}_{\ell}) - \tau_n (L_n + 1)((N_n + 1) B_n)^{L_n +1}} \Big) - \dfrac{C_3(2^j \rho) \big(n/(\log n )^2 \big) }{320 \mathcal{K}_{\ell}(F + \mathcal{K}) } \Bigg).
\end{align}
As in the proof of Theorem \ref{thm1}, considering \textbf{Case 1}: $\rho > 1$; \textbf{Case 2}: $1/n < \rho < 1$; \textbf{Case 3}: $0 < \rho < 1/n < 1$, we obtain for some $n_0 \in \N$ and for all $n >n_0$,
\begin{align}\label{Bound_B1n_2}
  \E [B_{1, n}] &\leq \int_ 0^ {\infty} P(B_ {1, n} > \rho) d\rho  \lesssim \dfrac{(\log  n) ^2}{n}.
\end{align}

\medskip

\noindent 
\textbf{Step 2: Bounding the second term in the right-hand side of (\ref{proof_excess_risk_v1_2})}.
Consider $h_{n}^{\diamond} \in \mathcal{H}_{\sigma, n} \coloneqq \mathcal{H}_{\sigma}(L_n, N_n, B_n, F)$ such that  
\begin{equation*}
R (h_{n}^{\diamond})  + J_n(h_{n}^{\diamond}) \leq \underset{h \in \mathcal{H}_{\sigma, n}} {\inf} \left[ R(h)  + J_n(h) \right] + \dfrac{1}{n}.
\end{equation*}
By going as in Step 2 in the proof of Theorem \ref{thm1}, we get
\begin{equation}\label{B_2_n_inf_2}
\E[B_{2, n}] \leq 2J_n(h_n^{\diamond}) + 2\mathcal{E}_{Z_0}(h_n^{\diamond}) \leq 2 \underset{h \in \mathcal{H}_{\sigma, n}}{\inf}\Big[ \mathcal{E}_{Z_0}(h) + J_n(h) \Big] + \dfrac{1}{n}.
\end{equation}
\medskip

\noindent
\textbf{Step 3: Bounding the expected excess risk}. 
The expected excess risk of $\widehat{h}_n$ is (see (\ref{proof_excess_risk_v1_2})):
\[ \E\big[\mathcal{E}_{Z_0}(\widehat{h}_n) \big] = \E[R(\widehat{h}_n) - R(h^{*}) ] = \E[B_{1, n}] + \E[B_{2, n}].
\]
So, from (\ref{Bound_B1n_2}) and (\ref{B_2_n_inf_2}), we have, for all $n > n_0$,
\begin{align*} %\label{proof expected_excess_risk}
\E\big[\mathcal{E}_{Z_0}(\widehat{h}_n) \big] 
  \leq 2 \Big(\underset{h \in \mathcal{H}_{\sigma}(L_n, N_n, B_n, F)} {\inf} \Big[ \mathcal{E}_{Z_0} (h)  + J_n(h) \Big]  \Big)  + \dfrac{\Xi (\log n) ^2}{n},
\end{align*}
for some constant $\Xi >0$.
Which completes the proof  of the Theorem \ref{oracle_inequality_expo_strong_mixing}.
\qed
\subsection{Proof of Corollary \ref{corol1}}
\begin{enumerate}
\item Let $L_n, N_n, B_n, F >0$ satisfying the conditions in Corollary \ref{corol1} and $h^{*} \in  \mathcal{C}^{s}(\mathcal{X}, \mathcal{K})$ with $s, \mk >0$.
Set $  \epsilon_n = \dfrac{ 1}{ \big( n^{(\alpha)} \big)^{\frac{s}{\kappa s + d} } } $ for some $\kappa \ge 1$.
From Theorem $3.2$ in \cite{kengne2025excess}, there exists a positive constants $L_0, N_0, B_0, S_0 >0$ such that with 
$L_n = \dfrac{s L_0}{\kappa s + d} \log\big(n^{(\alpha)} \big), N_n = N_0 \big(n^{(\alpha)} \big)^{\frac{d}{ \kappa s + d}}, S_n = \frac{s S_0}{\kappa s + d} (n^{(\alpha)})^{\frac{d}{\kappa s + d}}\log\big(n^{(\alpha)} \big), 
$
$ B_n = B_0 \big(n^{(\alpha)} \big)^{\frac{4(d + s)}{\kappa s + d}}$, there is a neural network $h_{n}^{\circ} \in \mathcal{H}_{\sigma, n} := \mathcal{H}_{\sigma}(L_n, N_n, B_n, F)$ satisfying,
\begin{equation}\label{cond_excess_risk_bound}
\| h_{n}^{\circ} - h^{*} \|_{\infty, \mathcal{X}} \leq \epsilon_n= \dfrac{ 1}{ \big( n^{(\alpha)} \big)^{\frac{s}{\kappa s + d} } }.
\end{equation}
Set
\[  \widetilde{\mathcal{H} }_{\sigma, n} := \{h \in  \mathcal{H}_{\sigma,n}, ~ \|h - h^*\|_{\infty, \mx}  \leq \epsilon_n\}.
\]
In addition, for any $h \in  \mathcal{H}_{\sigma,n}$, $J_n(h) \leq \lambda_n S_n$ where $\lambda \asymp \dfrac{ \big(\log \big(n^{(\alpha)} \big) \big) ^{\nu_3}}{n^{(\alpha)}}$ with $\nu_3 > 0$. So, for $n \geq \varepsilon_0^{-( \kappa + d/s )}$, under the assumption \textbf{(A4)}, from Theorem \ref{thm1}, and (\ref{cond_excess_risk_bound}), for all $h^{*} \in  \mathcal{C}^{s}(\mathcal{X}, \mathcal{K})$, we get,
\begin{align}
\nonumber \E[\mathcal{E}_{Z_0}(\widehat{h}_n) ] & \lesssim   \underset{h \in \mathcal{H}_{\sigma}(L_n, B_n, N_n, F)}{\inf}\Big[ \mathcal{E}_{Z_0}(h) + J_n(h) \Big]   +  \dfrac{1}{n^{(\alpha)}}   \\
\nonumber & \lesssim   \underset{h\in \mathcal{H}_{\sigma, n}}{\inf}\Big[\mathcal{E}_{Z_0}(h) + \lambda_n S_n\Big] + \dfrac{1}{n^{(\alpha)}} \\
\nonumber & \lesssim   \underset{h\in \widetilde{\mathcal{H} }_{\sigma, n}}{\inf}\|h - h^{*} \|_{\kappa, P_{X_0}}^{\kappa}  + \lambda_n S_n + \dfrac{1}{n^{(\alpha)}} 
\\
\nonumber & \lesssim   \underset{h\in \widetilde{\mathcal{H} }_{\sigma, n}}{\inf}\|h - h^{*} \|_{\infty, \mx}^{\kappa}  + \lambda_n S_n + \dfrac{1}{n^{(\alpha)}}  
\\
\nonumber & \lesssim  \big(n^{(\alpha)}\big)^{-\frac{\kappa s}{\kappa s + d}} +  \big(n^{(\alpha)}\big)^{\frac{d}{\kappa s + d} - 1} \big(\log\big(n^{(\alpha)}\big) \big)^{\nu_3 + 1}  + \dfrac{1}{n^{(\alpha)}} \lesssim \dfrac{ \log^{\nu_3 +1} \big(n^{(\alpha)} \big) }{ \big(n^{(\alpha)}\big)^{\frac{\kappa s}{\kappa s + d}} }.
\end{align}
Thus, 
\[  \underset{h^{*} \in \mathcal{C}^{s}(\mathcal{X}, \mathcal{K})}{\sup}\E[\mathcal{E}_{Z_0}(\widehat{h}_n) ] \lesssim \dfrac{ \log^{\nu} \big(n^{(\alpha)} \big) }{ \big(n^{(\alpha)}\big)^{\frac{\kappa s}{\kappa s + d}} }, \]
for all $\nu >3$.
\item  Following similar steps and arguments as in the first point of the proof of Corollary \ref{corol1}, with $\epsilon_n = \dfrac{1}{n^{\frac{s}{\kappa s + d}}}$, where $\kappa \geq 1$, $s > 0$ (the smoothness parameter of the Hölder space), and $d > 0$ (the dimension of the input space), we obtain:
\[  \underset{h^{*} \in \mathcal{C}^{s}(\mathcal{X}, \mathcal{K})}{\sup}\E[\mathcal{E}_{Z_0}(\widehat{h}_n) ] \lesssim \dfrac{ \big (\log n \big) ^{\nu}  }{ n^{\frac{\kappa s}{\kappa s + d}} }, \]
for all $\nu > 5$. \qed
\end{enumerate}
\subsection{Proof of Corollary \ref{corol01_v1}}
\begin{enumerate}
\item Let $L_n, N_n, B_n, F_n = F$ satisfying the conditions in Corollary \ref{corol01_v1}.
Choose $S_n \asymp  n^{(\alpha)} \phi_{n^{(\alpha)}} \log n^{(\alpha)}$ and set  $\mathcal{H}_{\sigma, n} \coloneqq \mathcal{H}_{\sigma}(L_n, N_n,, B_n, F_n, S_n)$.
Consider the class of composition structured functions $\mathcal{G}(q, \bold{d}, \bold{t}, \boldsymbol{\beta}, \mathcal{K})$ and let $h ^*
\in \mathcal{G}(q, \bold{d}, \bold{t}, \boldsymbol{\beta}, \mathcal{K})$.
From the proof of Theorem $1$ in \cite{schmidt2020nonparametric}, there exists a constant $C_0>0$ independent of $n$ such that, 
\begin{equation}\label{cond_excess_risk_bound_corol01_v1}
\underset{h^{*}\in \mathcal{G}(q, \bold{d}, \bold{t}, \boldsymbol{\beta}, A)}{\sup}~\underset{h \in \mathcal{H}_{\sigma, n}}{\inf}\|h - h^{*}\|_{\infty, \mathcal{X}}^2 \leq C_0 \underset{0\leq i \leq q}{\max}\big(n^{(\alpha)} \big)^{ \frac{-2\beta_i^* }{2\beta_i^* + t_i}   } = C_0 \phi_{n^{(\alpha)}}.  
\end{equation}
%
% 
%From Theorem \ref{thm1}, in addition to (\ref{cond_excess_risk_bound_corol01_v1}) and the fact that for any $h \in \mathcal{H}_{\sigma, n},
%
According the fact that for any $h \in \mathcal{H}_{\sigma, n}\|$, $J_n(h) \leq \lambda _n S _n$, where $\lambda_n \asymp \dfrac{\big(\log(n^{(\alpha)})\big)^{\nu_3}}{n^{(\alpha)}}$ with $\nu_3 > 2$. 
So, under \textbf{(A4)} and for sufficiently large $n$ such that $\sqrt{C _0 \phi_{n^{(\alpha)}} } \leq \varepsilon _0$, from Theorem \ref{thm1} and (\ref{cond_excess_risk_bound_corol01_v1}),
for all $h^{*} \in \mathcal{G}(q, \bold{d}, \bold{t}, \boldsymbol{\beta}, A)$, we get,
\begin{align}
\nonumber \E[\mathcal{E}_{Z_0}(\widehat{h}_n) ] & \leq 2 \Big(\underset{h \in \mathcal{H}_{\sigma} (L_n, N_n, B_n, F)}{\inf} \Big[\mathcal{E}_{Z_0}(h) + J_n(h) \Big] \Big) + \dfrac{ 1}{n^{(\alpha)}} \\
\nonumber & \leq 2 \Big(\underset{h \in \mathcal{H}_{\sigma,n} }{\inf} \Big[\mathcal{E}_{Z_0}(h) + J_n(h) \Big] \Big) +  \dfrac{1}{n^{(\alpha)}} 
\\
\nonumber & \leq 2  \Big(\underset{h \in \mathcal{H}_{\sigma,n} }{\inf} \Big[\|h- h^ {*} \|_{\infty, \mathcal{X}} ^{\kappa} + \lambda_ {n} S_n  \Big] \Big)  + \dfrac{1}{n^{(\alpha)}} 
\\
\nonumber & \lesssim \Big( \phi_{n^{(\alpha)}} ^{\kappa /2}  + \lambda_n S_n \Big)  +  \dfrac{1}{n^{(\alpha)}} \\
\nonumber & \lesssim  \phi_{n^{(\alpha)}} ^{\kappa/2}  + \phi_{n^{(\alpha)}} \big(\log\big(n^{(\alpha)} \big) \big)^{\nu_3 + 1}   +  \dfrac{1}{n^{(\alpha)}}   \\
\nonumber & \lesssim  \Big( \phi_{n^{(\alpha)}} ^{\kappa/2} \lor \phi_{n^{(\alpha)}} \lor \big( {n^{(\alpha)}} \big) ^{-1}\Big) \big(\log\big(n^{(\alpha)} \big) \big)^{\nu_3 + 1} \\
& \lesssim  \Big( \phi_{n^{(\alpha)}} ^{\kappa/2} \lor \phi_{n^{(\alpha)}} \Big) \big(\log\big(n^{(\alpha)} \big) \big)^{\nu_3 + 1}.
\end{align}
Therefore,
\[  \underset{h^{*} \in \mathcal{G}(q, \bold{d}, \bold{t}, \boldsymbol{\beta}, A)}{\sup} \E[\mathcal{E}_{Z_0}(\widehat{h}_n) ]   \lesssim  \Big( \phi_{n^{(\alpha)}} ^{\kappa/2} \lor \phi_{n^{(\alpha)}} \Big) \big(\log\big(n^{(\alpha)} \big) \big)^{\nu_3 + 1}, \]
for all $\nu_3 > 2$.
\item  Using similar steps and arguments as in the first point of the proof of Corollary \ref{corol01_v1}, with $S_n \asymp  n \phi_n \log (n)$, and $ \lambda_n \asymp \frac{(\log (n)) ^{\nu_3}}{n}$, where $\nu_3 > 2$, with the application of Theorem \ref{oracle_inequality_expo_strong_mixing}, we obtain,
\[   \underset{h^{*} \in \mathcal{G}(q, \bold{d}, \bold{t}, \boldsymbol{\beta}, A)}{\sup} \E[\mathcal{E}_{Z_0}(\widehat{h}_n) ]   \lesssim  \Big( \phi_{n} ^{\kappa/2} \lor \phi_{n} \Big) \big(\log(n) \big)^{\nu_3 + 1}, \]
for all $\nu > 5$. \qed
\end{enumerate}

\subsection{Proof of Lemma \ref{equa_lem1}}
The proof of this Lemma follows similar arguments from the proof of Theorem 3 in \cite{schmidt2020nonparametric}.

\medskip

\noindent
Let $i^{*} \in \underset{0\leq i \leq q}{\argmin}[\beta_{i^{*}}^{*}/(2\beta_{i^{*}}^{*} + t_i)]$. 
Therefore, $\phi_n = n^{-2\beta_{i^{*}}^{*}/(2\beta_{i^{*}}^{*} + t_{i^*})}$, where $\phi_n$ is defined in (\ref{def_phi_n}).
For simplicity, we set $\beta^{*} \coloneqq \beta_{i^{*}}$, $\beta^{**} \coloneqq \beta_{i^{*}}^{*}$ and $t^{*} \coloneqq t_{i^{*}}$.
Moreover, set $B = \prod_{\ell=i^* +1}^{q}(\beta_{\ell} \land 1)$.
So, we also have $\beta^{**} = \beta^{*}B$.
 Let $K \in L^2(\R) \cap \mathcal{C}_1^{\beta^{*}}(\R, 1)$ with support $[0, 1]$. 
%
%It is easy to see that such $K$ exists. 
%
Define $m_n \coloneqq \lfloor \rho n^{1/(2\beta^{**} + t^{*})}\rfloor$ and $h_n \coloneqq 1/m_n$ where $\rho$ is chosen such that $nh_n^{\beta^{**} + t^{*}} \leq \log(2)/(72 \|K^B\|_2^{t^{*}})$. 
Set
\[\mathcal{U}_n \coloneqq \{(u_1, \cdots, u_{t^{*}}): u_i\in \{0, h_n, 2h_n, \cdots, (m_n - 1)h_n\}\}.
\]
For any $\boldsymbol{u} = (u_1, \dots, u_{t^{*}}) \in \mathcal{U}_n$,
define,
\begin{equation}\label{equa1}
\psi_{\boldsymbol{u}}(x_1, \dots, x_{t^{*}}) \coloneqq h_n^{\beta^{*}}\prod_{j=1}^{t^{*}} K\big(\frac{x_j - u_j}{h_n} \big).   
\end{equation}
We have
\[ \|\psi_{\boldsymbol{u}}^B\|_2   = \Big[\displaystyle \int_{[0,1]^{t^*}} \Big( h_n^{\beta^{*}B}\prod_{j=1}^{t^{*}} K^B(\frac{x_j - u_j}{h_n})\Big)^{2} dx_1\dots dx_{t^{*}} \Big]^{1/2} 
  = \Big[\displaystyle\int_{ [0,1]^{t^*} } (h_n^{\beta^{*}B})^{2} \Big(\prod_{j=1}^{t^{*}} K^B(\frac{x_j - u_j}{h_n}) \Big)^{2} dx_1\dots dx_{t^{*}} \Big]^{1/2}. \]
Let $z_j = \frac{x_j - u_j}{h_n}$ for $j = 1, \dots, t^{*}$.
Since $K$ is supported on $[0,1]$, we have
\begin{align}\label{equa3}
\nonumber \|\psi_{\boldsymbol{u}}^{B}\|_2 & = \Big[\displaystyle\int_{ \big[ -\frac{u_1}{h_n}, \frac{1-u_1}{h_n} \big]\times\cdots\times \big[ -\frac{u_{t^*}}{h_n}, \frac{1-u_{t^*}}{h_n} \big] } (h_n^{\beta^{*}B})^{2} \Big(\prod_{j=1}^{t^{*}} K^{B}(z_j)\Big)^{2} h_n dz_1\dots h_n dz_{t^{*}} \Big]^{1/2} 
\\ 
 & = h_n^{\beta^{*}B + t^{*}/2} \Big[\displaystyle\int_{[0,1]^{t^{*}}} \Big(\prod_{j=1}^{t^{*}} K^B(z_j) \Big)^{2} dz_1\dots dz_{t^{*}} \Big]^{1/2}
  = h_n^{\beta^{**} + t^{*}/2}\|K^B\|_{2}^{t^{*}}.
\end{align}
Also, one can easily get that, for all $\boldsymbol{u}, \boldsymbol{u}' \in \mathcal{U}_n$ with $\boldsymbol{u} \ne \boldsymbol{u}'$,
\begin{equation}\label{proof_psi_u_u_prime_int_0}
 \int_{[0,1]^{t^*}} \psi_{\boldsymbol{u}}(x_1, \dots, x_{t^{*}}) \psi_{\boldsymbol{u}'}(x_1, \dots, x_{t^{*}}) dx_1\dots dx_{t^{*}} =0. 
\end{equation}
For any $\boldsymbol{W}=(\omega_{\boldsymbol{u}})_{\boldsymbol{u} \in \mathcal{U}_n} \in \{0,1\}^{|\mathcal{U}_n|}$, define for all $\boldsymbol{x} = (x_1,\cdots, x_{t^*}) \in [0,1]^{t^*}$,
\begin{equation}\label{equa6}
h_{\boldsymbol{W}}(\boldsymbol{x}) = \sum_{\boldsymbol{u} \in \mathcal{U}_n}\omega_{\boldsymbol{u}}\psi_{\boldsymbol{u}}(x_1,\dots, x_{t^{*}})^B.
\end{equation}
From (\ref{equa3}) and (\ref{proof_psi_u_u_prime_int_0}), we have (see also \cite{schmidt2020nonparametric}) for all $\boldsymbol{W}=(\omega_{\boldsymbol{u}})_{\boldsymbol{u} \in \mathcal{U}_n},  \boldsymbol{W}'=(\omega'_{\boldsymbol{u}})_{\boldsymbol{u} \in \mathcal{U}_n}  \in \{0,1\}^{|\mathcal{U}_n|}$,
\begin{align}\label{equa7}
\nonumber\|h_{\boldsymbol{W}} - h_{\boldsymbol{W'}}\|_2 & = \Big[\displaystyle\int_{\R^{t^{*}}} \Big( \sum_{\boldsymbol{u} \in \mathcal{U}_n}\omega_{\boldsymbol{u}}\psi_{\boldsymbol{u}}(x_1,\dots, x_{t^{*}})^B - \sum_{\boldsymbol{u} \in \mathcal{U}_n}\omega'_{\boldsymbol{u}}\psi_{\boldsymbol{u}}(x_1,\dots, x_{t^{*}})^B \Big)^2 dx_1\dots dx_{t^{*}} \Big]^{1/2}    
\\
\nonumber & = \Big[\sum_{\boldsymbol{u} \in \mathcal{U}_n} (\omega_{\boldsymbol{u}} -\omega'_{\boldsymbol{u}})^2\displaystyle\int_{\R^{t^{*}}} (\psi_{\boldsymbol{u}}(x_1,\dots, x_{t^{*}})^B)^2 dx_1\dots dx_{t^{*}} \Big]^{1/2} 
\\ 
 & = \sqrt{\text{Ham($\boldsymbol{W}$, $\boldsymbol{W'}$)}}\|\psi_{\boldsymbol{u}}^B\|_2 = \sqrt{\text{Ham($\boldsymbol{W}$, $\boldsymbol{W'}$)}} h_n^{\beta^{**} + t^{*}/2}\|K^B\|_2^{t^{*}}
\end{align}
where $\text{Ham($\boldsymbol{W}$, $\boldsymbol{W'}$)} = \sum_{\boldsymbol{u} \in \mathcal{U}_n}\ind_{\omega_{\boldsymbol{u}} \ne \omega'_{\boldsymbol{u}}}$.
From Lemma 2.9 in \cite{tsybakov2009nonparametric}, see also  the proof of Theorem 3 in \cite{schmidt2020nonparametric}), there exists a subset $\mathcal{W} \subset \{0,1\}^{m_n^{t^*}}$ satisfying $|\mathcal{W}| \geq 2^{m_n^{t^*}/8}$ such that $\sqrt{\text{Ham($\boldsymbol{w}$, $\boldsymbol{w'}$)}} \ge m_n^{t^{*}/2}/\sqrt{8}$ for all $\boldsymbol{W}, \boldsymbol{W}' \in \mathcal{W}$ with $\boldsymbol{W} \ne \boldsymbol{W}'$.
Recall that $m_n  = 1/h_n$.
According to (\ref{equa7}), we have for all $\boldsymbol{W}, \boldsymbol{W}' \in \mathcal{W}$ with $\boldsymbol{W} \ne \boldsymbol{W}'$, 
\begin{equation}\label{equa8}
 \|h_{\boldsymbol{W}} - h_{\boldsymbol{W'}}\|_2  \ge \frac{1}{\sqrt{8}}m_n^{t^{*}/2} h_n^{\beta^{**} + t^{*}/2}\|K^B\|_2^{t^{*}} 
 = \frac{1}{\sqrt{8}h_n^{t^{*}/2}} h_n^{\beta^{**} + t^{*}/2}\|K^B\|_2^{t^{*}} = \frac{1}{\sqrt{8}h_n^{t^{*}/2}} h_n^{\beta^{**} }\|K^B\|_2^{t^{*}}. 
\end{equation}
Let $\kappa = \|K^B\|_2^{t^{*}}/\sqrt{8 \rho^{\beta^{**}}}$ that implies $\|K^B\|_2^{t^{*}} = \kappa \sqrt{8\rho^{ \beta^{**}}}$.
Recall, $m_n = \lfloor \rho n^{1/(2\beta^{**} + t^{*})}\rfloor$ and $m_n = 1/h_n$, that is $\rho > \frac{n^{-2/(2\beta^{**} + t^{*})}}{h_n}$.
So, 
\begin{equation}\label{equa10}
\|K^B\|_2^{t^{*}}  = \kappa \sqrt{8} \sqrt{\rho^{\beta^{**}}} \ge \frac{\kappa \sqrt{8}}{h_n^{\beta^{**}}}\sqrt{n^{-2\beta^{**}/(2\beta^{**} + t^{*})}}.   
\end{equation}
Thus, (\ref{equa8}) and (\ref{equa10}) imply that, for all $\boldsymbol{W}, \boldsymbol{W}' \in \mathcal{W}$ with $\boldsymbol{W} \ne \boldsymbol{W}'$, 
\begin{equation}\label{equa11}
\|h_{\boldsymbol{W}} - h_{\boldsymbol{W'}}\|_2 \ge \kappa \sqrt{\phi_n},    
\end{equation}
with $ \phi_n = n^{-2\beta^{**}/(2\beta^{**} + t^{*})}$.
Moreover, from the proof of Theorem 3 in \cite{schmidt2020nonparametric}), $\widetilde{h}_{\boldsymbol{W}} \in \mathcal{G}(q, \bold{d}, \bold{t}, \boldsymbol{\beta}, \mathcal{K}) $ for sufficiently large $\mk$.
Thus, the item (i) of the lemma holds.

\medskip

 Furthermore, recall that the constant $\rho$ is chosen such that $n h_n^{\beta^{**} + t^{*}} \leq \log(2)/(72\|K^B\|_2^{t^{*}})$ and that $|\mathcal{W}| \geq 2^{m_n^{t^*}/8}$.
Since $|\mathcal{U}_n| = m_n^{t^*}$ and according to (\ref{equa8}), we have for all $\boldsymbol{W}, \boldsymbol{W}' \in \mathcal{W}$,
\begin{align}\label{equa12}
\nonumber n\|h_{\boldsymbol{W}} - h_{\boldsymbol{W'}}\|_2 &= n \sqrt{\text{Ham($\boldsymbol{W}$, $\boldsymbol{W'}$)}} h_n^{\beta^{**} + t^{*}/2}\|K^B\|_2^{t^{*}} \leq n m_n^{t^{*}/2} h_n^{\beta^{**} + t^{*}/2}\|K^B\|_2^{t^{*}}\\
& \leq n m_n^{t^{*}} h_n^{\beta^{**} + t^{*}}\|K^B\|_2^{t^{*}} \leq  \dfrac{ \log 2 }{72}  m_n^{t^{*}} \leq \dfrac{\log 2}{72} \times \dfrac{8 \log |\mathcal{W}| }{\log 2} \leq \frac{\log |\mathcal{W}|}{9}.  
\end{align}
This establishes the item (ii) and completes the proof of the lemma. \qed

\medskip

\subsection{Proof of Theorem \ref{thm2}}\label{Proof_thm2}
We will apply the Theorem 2.7 in \cite{tsybakov2009nonparametric}.
In this subsection, we set:
\begin{itemize}
\item  $\|\cdot\|_q = \|\cdot\|_{L^q[0,1]^d}$ for all $q\geq 1$;
%\item $\boldsymbol{Y}_n = (Y_n, \dots, Y_{-d+1})$;
\item For all $h^* \in \mathcal{M}(q, \bold{d}, \bold{t}, \boldsymbol{\beta}, A)$, $P_{h^*,n}$ denotes the distribution of $\boldsymbol{Y}_n = (Y_n, \dots, Y_{-d+1})$ where, $Y_i$, $i=-d+1,\cdots,n$ are generated from model (\ref{equa_model_auto});
\item The probability densities considered are  with respect to the Lebesgue measure;
\item For two probability measures $P_1$ and $P_2$, $KL(P_1, P_2)$ represents the Kullback-Leibler divergence of $P_1$ from $P_2$.  
\end{itemize}
Consider model (\ref{equa_model_auto}) with the error term that has a standard Laplace probability density.

\medskip

\noindent
By Lemma \ref{equa_lem1}, there is a constant $A > 0$ such that, for $n > 0$, there exist an integer $M \ge 1$ and function $h_{(0)}, \dots, h_{(M)} \in \mathcal{G}(q, \bold{d}, \bold{t}, \boldsymbol{\beta}, A)$  such that
\begin{equation*}
  \|h_{(j)} - h_{(k)}\|_{2} \ge \kappa \sqrt{\phi _n} ~ \text{for all} ~ 0\leq j < k \leq M,   
\end{equation*}
and,
\begin{equation}\label{proof_Laplace_appl_lem}
\frac{1}{M}\sum_{j=1}^M n\|h_{(j)} - h_{(0)}\|_2 \leq \frac{1}{9}\log(M) 
\end{equation}
with $\kappa >0$, independent of $n$.
So, $h_{(0)}^*=h_{(0)}\ind_{[0,1]^d}, \dots, h_{(M)}^*=h_{(M)}\ind_{[0,1]^d} \in \mathcal{M}(q, \bold{d}, \bold{t}, \boldsymbol{\beta}, A)$ and 
\[   \|h_{(j)}^* - h_{(k)}^* \|_{2}^2 \ge \kappa^2  \phi _n ~ \text{for all} ~ 0\leq j < k \leq M.\]
Thus the first item of Theorem 2.7 in \cite{tsybakov2009nonparametric} hold.

\medskip

\noindent Let us show the second item of Theorem 2.7 in \cite{tsybakov2009nonparametric}.  

\medskip

\noindent Let $P_j = P_{h_{(j)}^*, n}$ be the law of random vector $\boldsymbol{Y}_n = (Y_n, \dots, Y_{-d+1})$, where $Y_t$ with $(t = -d+1, \dots, n)$ are generated from model (\ref{equa_model_auto}) with standard Laplace error. 
For $t > d$, let $X_t = (Y_{t-1}, \dots, Y_{t-d})$.
For model (\ref{equa_model_auto}), the conditional density of $Y_t$ given $X_{t-d}$, is:
\begin{equation*}%\label{equa23}
g\big(y_t- h^*(X_{t-d}) \big) = \frac{1}{2}\exp(-|y_t- h^*(X_{t-d})|),   
\end{equation*}
where $ g(y) = \frac{1}{2}e^{-|y|}$ is the density of the standard Laplace distribution.
So, one can easily get,
\begin{equation*}%\label{equa24}
\frac{dP_j}{dP_0}(\boldsymbol{Y}_n) = \prod_{t=1}^n \frac{g(Y_t - h_{(j)}^*(Y_{t-1}, \dots, Y_{t-d}))}{g(Y_t - h_{(0)}^*(Y_{t-1}, \dots, Y_{t-d}))}. 
\end{equation*}
As a consequence,
\begin{align*}%\label{equa25}
\nonumber \E\Big[\log\frac{dP_j}{dP_0}(\boldsymbol{Y}_n) \Big] & = \sum_{t=1}^n \E[-|Y_t - h_{(j)}^*(X_t)| + |Y_t - h_{(0)}^*(X_t)|]
\\
\nonumber& \leq \sum_{t=1}^n \E[|h_{(j)}^*(X_{t-d}) - h_{(0)}^*(X_{t-d})|] \\
& \leq \sum_{t=1}^n \E[|h_{(j)}^*(X_{0}) - h_{(0)}^*(X_{0})|] \leq n \|\widetilde{h}_{(j)} - \widetilde{h}_{(0)}\|_{1}.
\end{align*}
Hence, the Kullback-Leibler divergence between of $P_j$ from $P_0$ satisfies,
\begin{equation*}%\label{equa28}
 \frac{1}{M}\sum_{j=1}^M KL(P_j, P_0) \leq n \|h_{(j)}^* - h_{(0)}^*\|_{1}.    
\end{equation*}
Therefore, by using the H\"older inequality and (\ref{proof_Laplace_appl_lem}), it holds that,
\begin{equation*}%\label{equa28}
   \frac{1}{M}\sum_{j=1}^M KL(P_j, P_0) \leq n \|h_{(j)}^* - h_{(0)}^*\|_{2} \leq \frac{1}{9}\log M.    
\end{equation*}
So, the second attempt in Theorem 2.7 in \cite{tsybakov2009nonparametric} is satisfied.
Thus, one can find a positive constant $ C_{\kappa}$ such that
\begin{equation*} 
\underset{\widehat{h}_n}{\inf}~\underset{h^*_0\in\mathcal{M}(q, \bold{d}, \bold{t}, \boldsymbol{\beta}, A)}{\sup} \E[\|\widehat{h}_n - h^*_0\|_{2}^2] \ge C_{\kappa}\phi _n.
\end{equation*}
This completes the proof of the theorem. 
\qed

\end{document}